\documentclass[10pt,twocolumn,letterpaper]{article}
\usepackage{cvpr}              % To produce the CAMERA-READY version
\usepackage{graphicx}
\usepackage{amsmath}
\usepackage{amssymb}
\usepackage{booktabs}
\usepackage[acronym]{glossaries}
\usepackage{mathtools} % amsmath with fixes and additions

\usepackage{tikz} % nice language for creating drawings and diagrams

\usepackage{latexsym}
\usepackage{amsthm}

\usepackage{amsfonts}

\usepackage{subcaption}

\usepackage{algorithm}
\usepackage{algorithmic}

\newcommand{\csize}{
\fontsize{8}{8}\selectfont
}

\newcommand{\mb}[1]{\mathbf{#1}}

\newcommand{\thmref}[1]{Theorem~\ref{#1}}
\newcommand{\tabref}[1]{Table~\ref{#1}}
\newcommand{\figref}[1]{Figure~\ref{#1}}
\newcommand{\eqnref}[1]{Eq.~\ref{#1}}
\newcommand{\secref}[1]{Section~\ref{#1}}

\newcommand{\algoref}[1]{Algorithm~\ref{#1}}

\newtheorem{theorem}{Theorem}

\newcommand{\beqa}{\begin{eqnarray}}
\newcommand{\eeqa}{\end{eqnarray} }
\newcommand{\beq}{\begin{equation}}
\newcommand{\eeq}{\end{equation} }

\newcommand{\beqan}{\begin{eqnarray*}}
\newcommand{\eeqan}{\end{eqnarray*} }
\newcommand{\beqn}{\begin{equation*}}
\newcommand{\eeqn}{\end{equation*} }

\newcommand{\R}{\mathbb{R}}

\definecolor{shadecolor}{gray}{0.975}
\newcommand{\algshade}[1]{
    \hspace*{-\fboxsep}
    \colorbox{shadecolor}{
        \parbox{\linewidth}{#1}
    }
}

\definecolor{lavenderblush}{rgb}{1.0, 0.97, 0.98}
\newcommand{\algredshade}[1]{
    \hspace*{-\fboxsep}
    \colorbox{lavenderblush}{
        \parbox{\linewidth}{#1}
    }
}

\definecolor{aliceblue}{rgb}{0.97, 0.985, 1.0}
\newcommand{\algblueshade}[1]{
    \hspace*{-\fboxsep}
    \colorbox{aliceblue}{
        \parbox{\linewidth}{#1}
    }
}

\renewcommand\algorithmiccomment[1]{
  {
  	{
  	{\textit{\%\ #1}}
  	}
  }
}

\definecolor{cvprblue}{rgb}{0.21,0.49,0.74}
\usepackage[pagebackref,breaklinks,colorlinks,citecolor=cvprblue]{hyperref}

\def\paperID{15912} % *** Enter the Paper ID here
\def\confName{CVPR}
\def\confYear{2024}

\title{Deep Generative Sampling in the Dual Divergence Space:\\
A Data-efficient \& Interpretative Approach for Generative AI}

\author{
Sahil Garg\textbf{*}, Anderson Schneider, Anant Raj, Kashif Rasul, Yuriy Nevmyvaka\\
Dept. of Machine Learning Research, Morgan Stanley\\
{
\textbf{*}Corresponding Author: 
sahil.garg@morganstanley.com, sahil.garg.cs@gmail.com
}
\and
Sneihil Gopal\\
PREP Associate, NIST and Dept. of Physics, Georgetown University\\
\and
Amit Dhurandhar, Guillermo Cecchi\\
IBM Research\\
\and
Irina Rish\\
Mila - Quebec AI Institute and Université de Montréal\\
}

\setacronymstyle{long-short}
\newacronym{ID}{ID}{in-distribution}
\newacronym{VAE}{VAEs}{Variational Autoencoders}
\newacronym{NF}{NFs}{Normalizing Flow}
\newacronym{DDPM}{DDPMs}{Denoising Diffusion Probabilistic Models}
\newacronym{MVT}{MVT}{multivariate time series}
\newacronym{ViT}{ViTs}{Vision Transformers}
\newacronym{EEG}{EEG}{electroencephalogram}
\newacronym{OOD}{OOD}{out-of-distribution}
\newacronym{GAN}{GANs}{Generative Adversarial Networks}
\newacronym{WGAN}{W-GANs}{Wasserstein Generative Adversarial Networks}
\newacronym{FGAN}{f-GANs}{f-GAN}
\newacronym{knn}{kNNs}{k-nearest neighbors}
\newacronym{CNN}{CNNs}{Convolutional Neural Networks}
\newacronym{resnet}{ResNets}{Residual Neural Networks}
\newacronym{FNN}{FNNs}{Feedforward Neural Networks}
\newacronym{FID}{FID}{Fréchet Inception Distance}
\newacronym{DNN}{DNN}{deep neural network}
\newacronym{MMI}{MMI}{multivariate mutual information}
\newacronym{SDE}{SDEs}{stochastic differential equation}
\begin{document}

\maketitle

\begin{abstract}
\vspace{-5mm}
Building on the remarkable achievements in generative sampling of natural images, we propose an innovative challenge, potentially overly ambitious, which involves generating samples of entire multivariate time series that resemble images. This would prove to be a valuable tool for professionals like neurologists, psychiatrists, environmentalists, and economists, among others.
However, the statistical challenge lies in the small sample size, sometimes consisting of a few hundred subjects. This issue is especially problematic for deep generative models that follow the conventional approach of generating samples from a standard distribution and then decoding or denoising them to match the true data distribution.
In contrast, our method is grounded in information theory and aims to implicitly characterize the distribution of images, particularly the (global and local) dependency structure between pixels. We achieve this by empirically estimating its KL-divergence in the dual form with respect to the respective marginal distribution. This enables us to perform generative sampling directly in the optimized one-dimensional dual divergence space. 
Specifically, in the dual divergence space, training samples representing the data distribution are embedded in the form of various clusters between two end points. In theory, any sample embedded between those two end points is \gls{ID} w.r.t. the data distribution.
Our key idea for generating novel samples of images is to interpolate between the clusters via a walk as per gradients of the dual function w.r.t. the data dimensions.
In addition to the data efficiency gained from direct sampling, we propose an algorithm that offers a significant reduction in sample complexity for estimating the divergence of the data distribution with respect to the marginal distribution.
We provide strong theoretical guarantees along with an extensive empirical evaluation using many real-world datasets from diverse domains, establishing the superiority of our approach w.r.t. state-of-the-art deep learning methods.
\end{abstract}

\section{Introduction}
    
\begin{figure}
\centering
\includegraphics[width=\columnwidth]{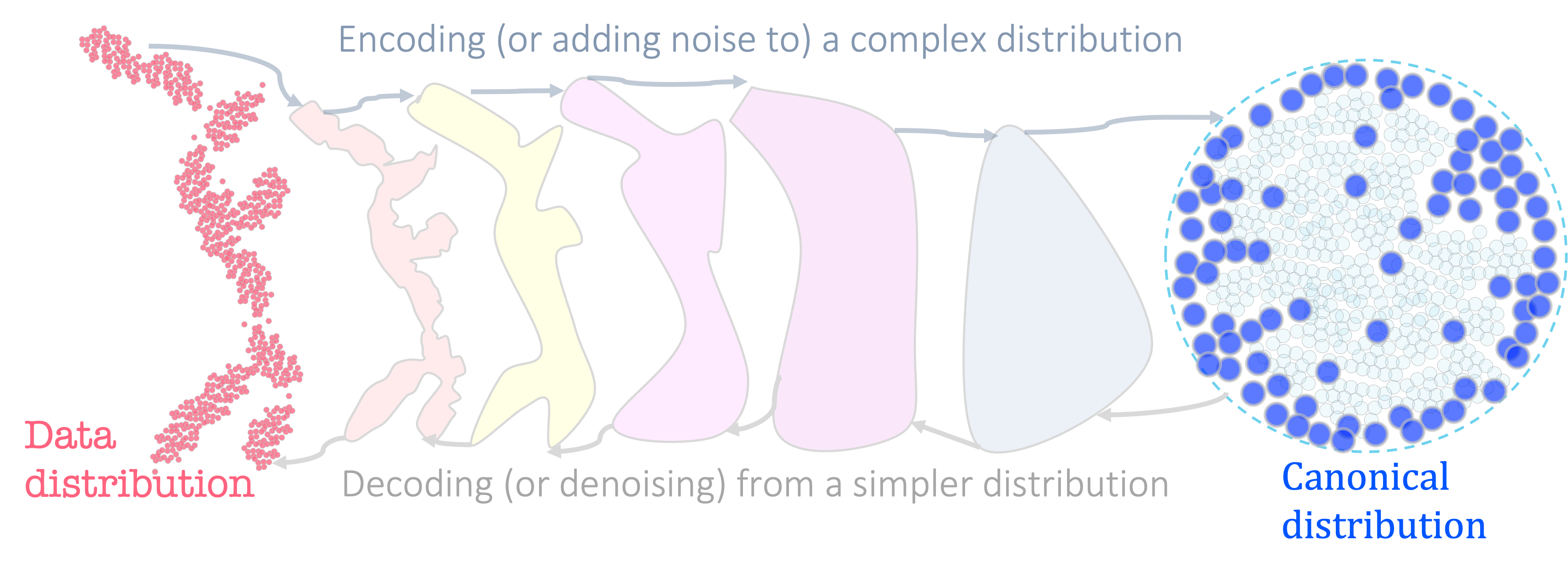}
\vspace{-2mm}
\caption{An illustration of the general paradigm followed by most approaches in the literature of deep generative sampling. The data distribution (represented by red dots) gradually evolves into a simpler canonical distribution, such as a Gaussian distribution, either through an encoder, as seen in \gls{VAE} and \gls{NF}, or by adding noise, as is the case in \gls{DDPM}. A canonical distribution facilitates the generation of novel samples which are then mapped back into the data distribution via a decoder as in \gls{VAE}, \gls{NF}, etc., or via a denoising diffusion process as in \gls{DDPM}. The intermediate distributions between the data distribution and the canonical distribution are implied in models such as \gls{VAE} while being explicit in \gls{NF} or \gls{DDPM}. One common limitation of these approaches is that they require a large sample size, which is not available for our problem.}
\label{fig:generative_sampling_paradigm}
\vspace{-5mm}
\end{figure}

Major advancements have been accomplished in the literature of generating natural images, especially owing to the success of denoising diffusion models~\cite{ho2020denoising,song2020score,song2020denoising,nichol2021improved,eschweiler2023denoising,khader2023denoising,zimmermann2021nested}.
One natural question that arises is whether we can translate this success story to other generative problems that have been of lesser interest to the wider community of computer vision. If so, this would present an opportunity to inherently change the way medical practitioners, such as psychiatrists and neurologists, would operate in the future.
Consider, for instance, analyzing \gls{EEG} signals for the entire duration across all the channels for a patient diagnosed with schizophrenia as if it's a single image. Generating samples of such images characterizing the missing schizophrenia patients can be highly valuable. Other such use cases include analyzing climate variables like pollution, wind or solar energy, city traffic, or stock markets, etc.

% \todo{cite} \todo{show such actual images or the general formatting of MVT as images}

% \begin{figure}
% \centering
% % 
% % \begin{subfigure}{0.5\textwidth}
% % \includegraphics[width=.99\linewidth]{mts_images/org_16.pdf}
% % \end{subfigure}
% % % 
% % \vspace{-3mm}
% % 
% \begin{subfigure}{0.5\textwidth}
% \includegraphics[width=\linewidth]{mts_images/org_17.pdf}
% \end{subfigure}
% % 
% % \vspace{-3mm}
% % 
% \begin{subfigure}{0.5\textwidth}
% \includegraphics[width=\linewidth]{mts_images/org_18.pdf}
% \end{subfigure}
% % 
% \caption{Air quality across major cities in the world monitored for every week as an image.}
% \label{fig:generative_sampling_entropy_div_wrt_data_dist}
% \end{figure}

To tackle this unique, unexplored problem of generating entire \gls{MVT} that resemble images, we can utilize state-of-the-art algorithms for generative sampling of images and highly expressive neural architectures, such as \gls{ViT}~\cite{dosovitskiy2020image}. However, this task also presents some novel challenges. For instance, considering the use case of \gls{EEG} recordings from patients suffering from schizophrenia, the sample size, or the number of patients for whom recordings can be obtained, is extremely limited, typically consisting of a few hundred patients or even fewer.
For an optimal use of such a small number of data points in high dimensions, especially in the context of deep learning, we introduce a fundamentally new approach that is rooted in information theory.
    
\begin{figure}
\centering
\includegraphics[width=\columnwidth]{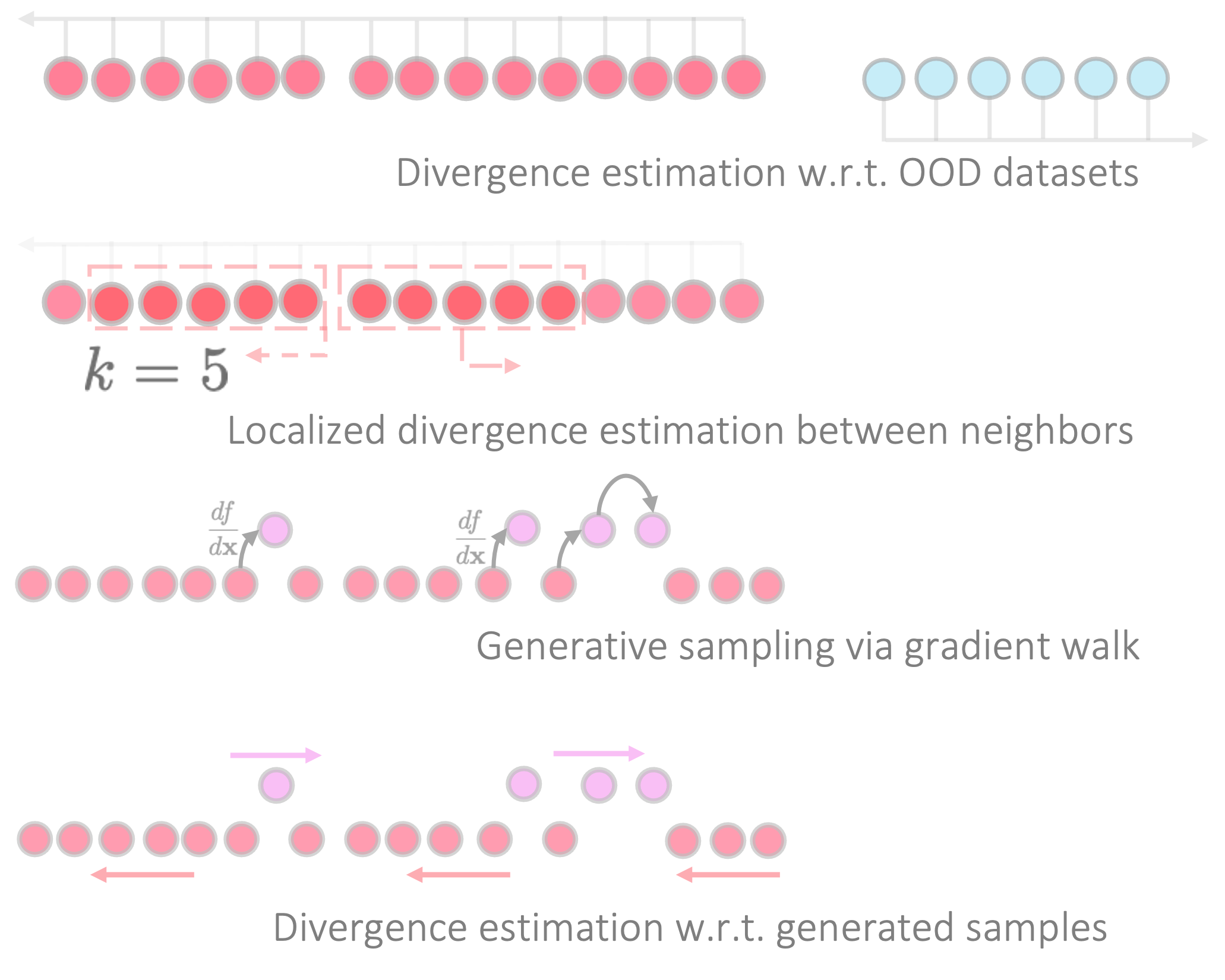}
\caption{A high-level illustration of our approach for generative sampling. Our key idea is to estimate empirical divergence in its dual form between the observed data points~(red dots) and the respective \gls{OOD} samples~(blue dots), so as to implicitly characterize the data distribution of interest in the 1-D dual functional space. 
The top sketch shows, in the dual space, samples from the two distributions~(red vs blue) that are pulled in opposite directions to attain the maximal estimate of the divergence as the optimal measure.
The boundary of real samples in the dual space (implicitly) represents the data distribution.
% 
% We generate samples as per the gradient of the dual function w.r.t. the inputs.
% 
% Besides the divergence estimation w.r.t. OOD samples, 
% 
For a finer-grained representation of real samples in the dual space, we estimate divergence locally between the nearest neighboring sets. Since the dual space is one-dimensional, it is highly interpretable and straightforward to identify regions (holes) of missing data points. Our algorithm generates (missing) samples in those holes via a gradient walk between the respective clusters. For robustness in generative sampling, we estimate divergence of the observations w.r.t. the generated samples, locally as well as globally.}
\label{fig:dv_generative_sampling}
\end{figure}
        
To understand the motivation behind our novel contributions to generative sampling, we first discuss the general paradigm behind state-of-the-art deep learning approaches for generative sampling, shown in \figref{fig:generative_sampling_paradigm}. A shared aspect of some of these methods is their reliance on a highly expressive neural decoder as in \gls{VAE}~\cite{kingma2013auto}, \gls{GAN}~\cite{goodfellow2014generative,nowozin2016f,nock2017f,arjovsky2017wasserstein,gulrajani2017improved,gemici2018primal}, or \gls{NF}~\cite{papamakarios2017masked,chen2018continuous}, among others. Another well-established family of score-based methods like \gls{DDPM}~\cite{ho2020denoising,song2020score}, maps samples generated usually from a canonical \emph{base} distribution~(typically Gaussian) to the data distribution by learning a highly expressive score. In other approaches like f-GANs and Wasserstein GANs, samples are generated such that (empirical) divergence between (the distributions of) generated and observed samples is minimized. Naturally, for training such decoders or denoising processes, one needs a reasonably large sample size, at least many thousands, as typically available for general-purpose natural images. For small datasets, one can imagine that the such machinery is bound to fail due to overfitting. % (along with heavy fine tuning over neural architectures as it's been done over the decades). Otherwise, with sample size in hundreds as available for our problem, overfitting is inevitable.
    
\begin{figure}
\centering
\includegraphics[width=\columnwidth]{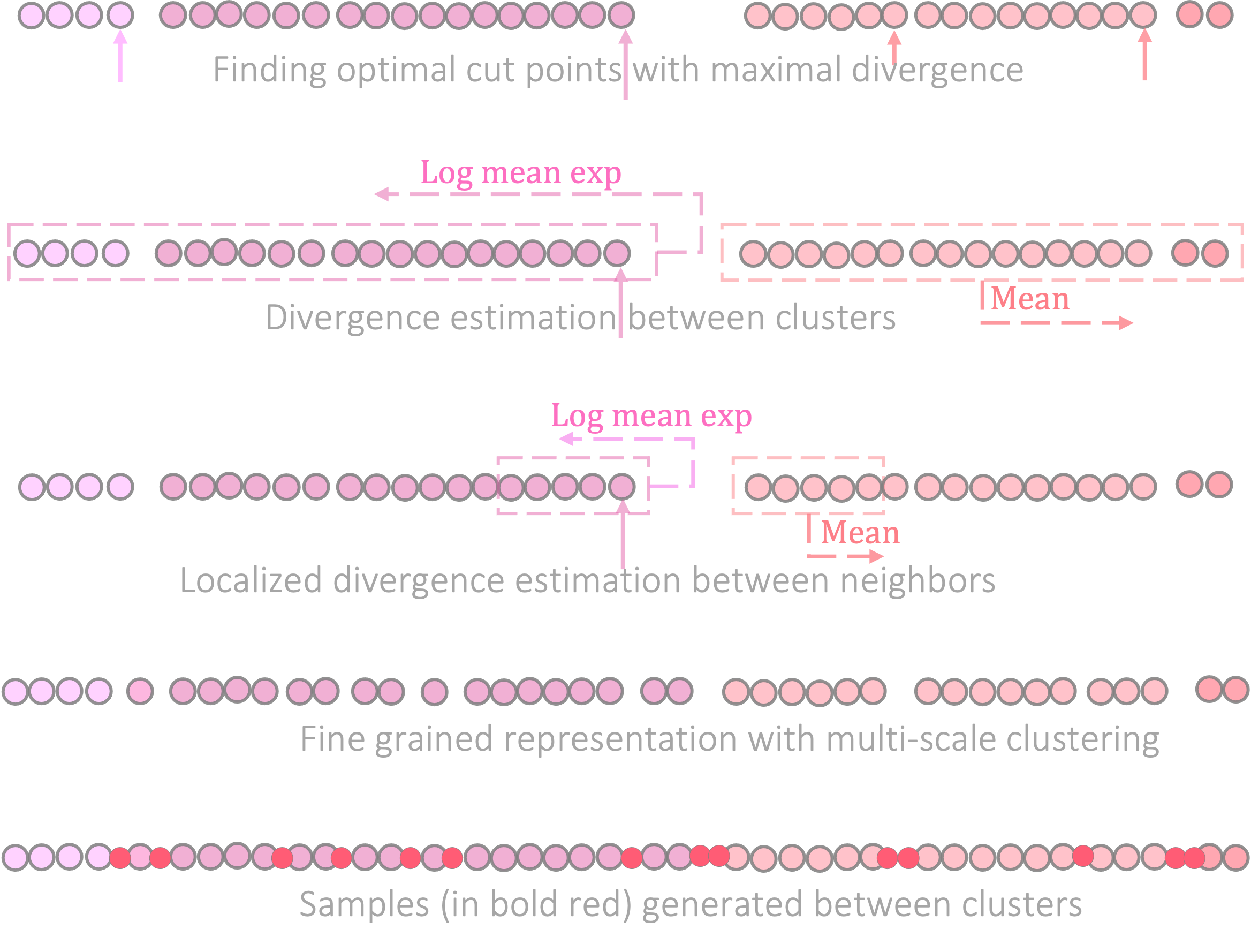}
\caption{An illustration of our approach for localized divergence estimation at cut points for multi-scale clustering. As established in \cite{gargitc23}, clusters with maximal (empirical) divergence w.r.t. each other are contiguous in the dual space separated by cut points. As such, the divergence between two clusters at a cut point is estimated by computing softmax and mean statistics using all the samples from the clusters respectively. We instead propose to estimate divergence locally at a cut point by computing the statistics only on the nearest neighbors of the cut point from either side, as shown above. By maximizing localized divergence between neighbors for a small number of cut points while minimizing it on average, we accomplish multi-scale clustering. Having optimized a fine-grained representation of real samples in the dual space, novel samples can be generated in empty space between the clusters~(or between distant neighboring data points).
% 
% \todo{refine and expand, explain cut points especially}
% 
}
\label{fig:localized_div_est_knn}
\end{figure}

In consideration of the above, owing to the recent advancements at the intersection of information theory and deep learning, we posit that it is not necessary to sample from a base (canonical) distribution for generating sampling. Instead, it is possible to generate novel samples directly from the one-dimensional dual space of the data distribution which can be obtained by estimating its empirical divergence in the dual form~\cite{donsker1983asymptotic} w.r.t. samples from a base canonical distribution (or for that matter, any set of samples deemed as out of distribution). We illustrate this idea in \figref{fig:dv_generative_sampling}. 
In reference to the figure, besides estimating the dual divergence of the data distribution w.r.t. base (canonical) distribution for representing real samples globally in the dual space, we propose to estimate divergence locally between nearest neighbors within the data distribution~(defined as per the global representation of inputs in the dual space itself) so as to learn a fine-grained representation of inputs. We refer the reader to \figref{fig:localized_div_est_knn} for a more detailed illustration of this particular idea which we refer to as ``localized divergence estimation between \gls{knn} for multi-scale clustering". 
Given a representation of real samples in the dual space, as shown in \figref{fig:dv_generative_sampling}, we propose to generate novel samples simply by a (gradient) walk in the dual space between real samples. To ensure robustness in generating samples that are \gls{ID} w.r.t. data distribution, we estimate empirical divergence between the two as well. 
Overall, all the steps mentioned above are performed in an iterative manner for learning the (neural) divergence estimator. Note, although in theory there is a uniquely optimal dual function for estimating divergence between two distributions, we advocate that, in practice, a single expressive neural encoder of images can be universally applied for estimating divergence between different distributions as well as at different scales~(local vs global).

\begin{figure}[t!]
\centering
\includegraphics[width=\columnwidth]{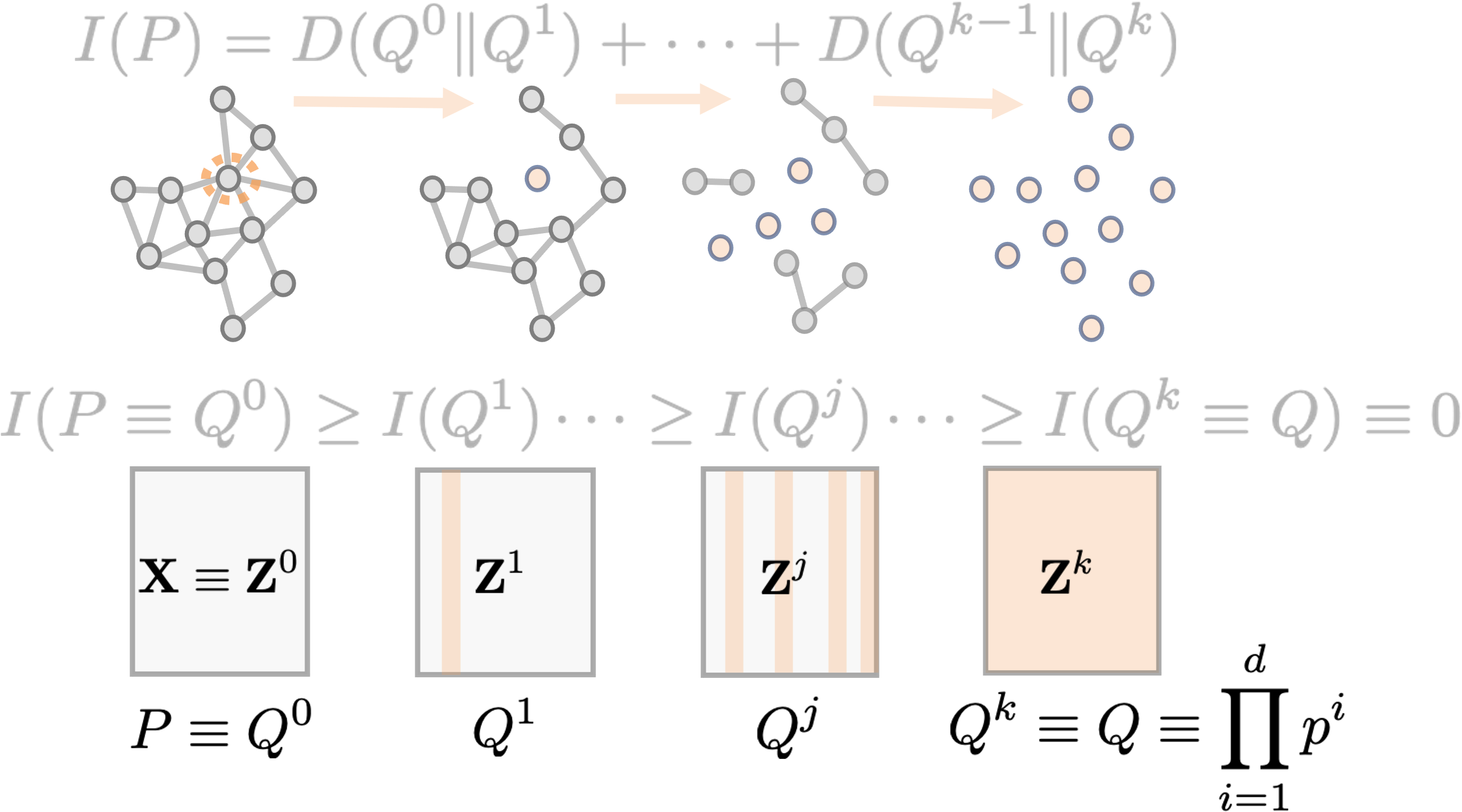}
\caption{An illustration for sample efficient divergence estimation w.r.t. marginals via dependency diffusion. 
For an input image, there is an underlying dependency structure between all the dimensions~(pixels) which is unknown and never learned explicitly.
From the left to right, in each step, we choose a subset of columns in the image and replace the values in there with samples from the respective marginal densities. This, in essence, diffuses the dependencies of the subset of the columns w.r.t. each other (as well as between pixels within each column) and w.r.t. all the other columns as shown in the dependency graphs. In this manner, we obtain samples for all the intermediate densities as well as the marginal distribution without having to learn the density functions or the respective dependency graphs. 
For each pair of adjacent distributions on the path,  $Q^{j-1}$ and $Q^j$, we obtain an empirical estimate of $D(Q^{j-1} \| Q^j)$ in its dual form, and the sum of all the divergence measures along the path gives us an estimate of divergence of the data distribution w.r.t. the marginals, $D(P \| Q)$. The estimation via the path avoids the otherwise exponential sample complexity w.r.t. true measure of $D(P \| Q)$~(in LB by \cite{song2019understanding}).
% 
% \todo{cut it short, explain specifically from perspective of modeling (may be MVT as) images.}
}
\label{fig:diffuse_dependencies}
\end{figure}

Furthermore, while the proposed approach is generic, we choose to estimate divergence of data distribution specifically w.r.t. the respective marginal distribution (as base distribution) to implicitly model the underlying dependencies between pixels in images, such as those corresponding to neural co-activations in the brain. We argue that it is a particularly suitable choice for our problem of modeling \gls{MVT} as images, unlike natural images, where it may not suffice to simply have an inductive bias of modeling dependency structure through a choice of neural architecture, such as deep convolutional networks, neither is it practically feasible to learn the dependency structure explicitly, without making any assumptions about it. Moreover, divergence w.r.t. marginals can be estimated with low sample complexity as we propose in this paper; see \figref{fig:diffuse_dependencies} for details.
    
\paragraph{Contributions}\footnote{Note that the supplementary material including the code base will be released only upon the publication of this article.} Our contributions are: (i) We introduce the problem of generative sampling of \gls{MVT} as an image, and conduct an extensive experimental study, using state-of-the-art deep generative models, for many real-world datasets including \gls{EEG} recording from patients diagnosed with schizophrenia, spiking neural activity in mouse brains, solar or wind energy, electrical consumption, traffic, pollution, stock returns, etc. (ii) Considering that low sample size is a bottleneck for the success of present methods for the proposed problem, we introduce a novel approach for generative sampling leveraging recent advancements in the literature of divergence estimation via deep learning. Our approach is equipped with efficient algorithms with low sample complexity, and built upon core concepts as introduced above which are simple, intuitive, yet highly effective. (iii) We provide information-theoretic guarantees and demonstrate the empirical superiority of our approach w.r.t. the well-known baselines in the literature of deep generative sampling.

% \paragraph{Other Related Works}
% \todo{a short para on connections to score based models, and fGANS, wGANS.}
    
% \todo{motivate divergence estimation w.r.t. marginals, why it is especially relevant for MVT as images.}
        
% For a given training set of data points~(MVT as images in our problem), we correspondingly obtain a set of data points deemed as OOD w.r.t. the training set. This OOD set can itself be another set of images or it can be synthetic samples from a canonical distribution such as a marginal, Gaussian, Uniform, etc. Optionally, in spirit of DDMs like models as shown in \figref{fig:generative_sampling_paradigm}, one can utilize samples from the intermediate distributions between the two extreme ends as additional OOD samples in our approach. 
    
% \newpage

% \subsection{Other Related Works}
    
% While this approach of ours completely avoids learning a decoder from a canonical distribution to data distribution, it \emph{does} need estimating divergence between the two. 

% \todo{score based models, fGANs, etc.}

\section{Deep Generative Sampling in the Dual Space}

% Our overall apprach is 

% \todo{give high level intuitive view of our approach and the subsections}
    
Let $X \in \R^{d \times d}$ be a random variable corresponding to (unknown) data distribution $P$ of images, and let $Q$ be the respective distribution for the product of the marginals \footnote{We assume the number of rows and columns to be same in images for an ease of understanding.}. %just as a notational convenience.}. 
We are given as input a set of real samples $\mb{X}$, representative of the data distribution. Our objective is to generate a set of novel samples, $\mb{X}^g$, from the data distribution. 

First, we introduce core concepts as the basis of our approach for generative sampling, i.e. (sample efficient) divergence estimation of data distribution w.r.t. the marginals in \secref{sec:div_wrt_marginals}, and local divergence estimation between \gls{knn} in the dual space in \secref{sec:local_div_est_knn}. Finally, we present the overall algorithm for generating samples in \secref{sec:overall_algo_sampling}. 
    
\subsection{Dual Divergence Estimation w.r.t. Marginals}
\label{sec:div_wrt_marginals}
We propose to characterize the data distribution by estimating KL-divergence of $Q$ w.r.t. $P$ in its dual form~\cite{donsker1983asymptotic} as,    
\begin{align}
D( Q \| P)
=
\max_{f(.) \in L^{\infty}}
\mathbb{E}_{\mb{z} \sim Q} \ f(\mb{z})
-
\log \mathbb{E}_{\mb{x} \sim P}\  e^{f(\mb{x})}.
\end{align}
Here, $f(.)$, is any function from the space of locally $\infty$-integrable functions such that expectations in the expression are finite, referred to as the \emph{dual function}. Correspondingly, the empirical divergence between samples $\mb{X}$ from data distribution $P$ and $\mb{Z}$ from marginal distribution $Q$ is defined as,
\begin{align}
\hat{D}(\mb{Z} \| \mb{X})
=
\max_{\hat{f}(.)\in L^{\infty}}
\sum_{\mathbf{z} \in \mb{Z}}
\frac{\hat{f}(\mb{z})}{m}
-
\log
\sum_{\mathbf{x} \in \mb{X}}
\frac{e^{\hat{f}(\mb{x})}}{n}
\label{eq:dual_div_eq}
\end{align}
For the empirical estimate of divergence, the dual divergence function $\hat{f}(.)$ can be a \emph{deep neural net}~\citep{belghazi2018mutual}. For dual representation of images, we can employ any vision architecture such as ViTs~\cite{dosovitskiy2020image}, CNNs, ResNets~\cite{targ2016resnet}, etc. For stable learning of the dual function with deep neural nets, standard implementation details such as early stopping, large batch size, low learning rate, gradient clipping, exponential moving averages, are known to be effective in practice~\citep{song2019understanding}.
    
\paragraph{Justifying Marginal as the Base Distribution}
The key advantage of taking the product of marginals as the base distribution is its information-theoretic interpretation: the divergence of data distribution w.r.t. the marginals represents the multivariate mutual information among the pixels within an image. This choice of base distribution is particularly suitable for domains such as neuroscience, healthcare, or finance, where marginal distributions are heavily long-tailed and have unique (unknown) dependency structures across input dimensions~\cite{breakspear2017dynamic,roberts2015heavy,mandelbrot1997variation,takada2001nonparametric,harvey2012kernel}. 

% For instance, neural activity data is heavy-tailed and highly non-Gaussian   
% 
% For instance, some subsets of neurons in a human brain co-activate rarely in contrast to others.
% 
% Marginal distributions of price changes of stocks can be significantly and arbitrarily different from Gaussian distributions \citep{mandelbrot1997variation, fama1965behavior}, often heavy tailed and asymmetric, which makes the problem of characterizing the data distribution challenging \citep{, harvey2012kernel}.
% 
% further aggravated due to the structural differences between the distribution of liquid vs relatively non-liquid securities
% 
% Estimation of joint density of various securities is . 
% 
% Thus, the only information that is reliably preserved is that shared between the data distribution under consideration and the base distribution is the marginal densities, corresponding to lower \kld between the two. 

\begin{algorithm}[tp!]
\caption{Estimate divergence locally between kNNs}
\begin{algorithmic}[1]
\REQUIRE{$\mb{X}$, $f(.)$}\\
\STATE $\mb{f} \gets$ f($\mb{X}$);\ \ $\mb{f} \gets$ sort($\mb{f}$)
% \COMMENT{estimate the dual function for inputs}
% \STATE $\mb{f} \gets$ sort($\mb{f}$)
% \COMMENT{sort inputs in the dual space}
\FOR{$j=0 \to n$}
\STATE $\mb{f}_{lnn} \gets \mb{f}[j-k:j]$ \COMMENT{kNNs from l.h.s.}
\STATE $\mb{f}_{lnn} \gets \mb{f}[j:j+k]$ \COMMENT{kNNs from r.h.s.}
\STATE $\mb{d}_{knn}[j] \gets$ D($\mb{f}_{lnn}$, $\mb{f}_{rnn}$)
\COMMENT{local divergence}
\ENDFOR
\STATE \textbf{Return} $\mb{d}_{knn}$
\end{algorithmic}
\label{alg:loc_div_est}
\end{algorithm}
    
\subsubsection{Sample Efficient Divergence Estimation via Path}
\label{sec:dependency_diffusion_path}
    
As we illustrate in \figref{fig:diffuse_dependencies}, rather than a direct estimate of divergence between $P$ and $Q$, we propose to estimate it indirectly by establishing a path of diffusing intra- and inter-column dependencies in images towards the base density $Q$.
\begin{align}
% &
P \equiv Q^0
\to
\cdots
\to
Q^j
\to
\cdots
\to
Q^{k} \equiv Q
% \equiv p^1 \cdots p^d
% \nonumber
% 
\\
\mb{X} \equiv \mb{Z}^0
\to
\mb{Z}^1
\to
\cdots
\to
\mb{Z}^j
\to
\cdots
\to
\mb{Z}^{k}
\equiv
\mb{Z}
\nonumber
% 
% \\
% Q^j(\mb{x})
% \equiv
% p^1 p^2 \cdots p^j
% P(x^{j+1}, \cdots, x^d)
% \nonumber
\\
I(P)
\geq
\cdots
\geq
I(Q^j)
\geq
\cdots
\geq
I(Q)
\equiv 0
\nonumber
\label{eqn:diffusion_path}
\end{align}
Here, $I(.)$ is the multi-variate mutual information function, characterizing dependencies between all the pixels in an image. Our key insight driving the reasoning above is that the divergence of the data distribution w.r.t the product of marginals, $D(P \| Q)$, is equal to the sum of divergence between adjacent distributions along the path of dependency diffusion~\cite{watanabe1960information}. The same logic applies to estimating $D( Q\| P )$.
\begin{align}
% 
% I(P)
% \equiv
D( P \| Q)
\equiv
D(Q^0 \| Q^1)
+
\cdots
+
D(Q^{k-1} \| Q^{k})
% \nonumber
% 
\\
D(Q \| P)
\equiv
D(Q^{k} \| Q^{k-1})
+
\cdots
+
D(Q^1 \| P)
\nonumber
\end{align}
Note that there is no need to explicitly learn the intermediate distributions, $Q^1, Q^2, \cdots, Q^{k-1}$, and neither do we need to sample directly from those distributions. We only need to learn the marginal distribution $Q$ which is a tractable problem, from which we take samples $\mb{Z}$. Given $\mb{X}$ and $\mb{Z}$, obtaining samples for the intermediate distributions is straightforward. Thus, we estimate $\hat{D}(\mb{Z} \| \mb{X})$ indirectly via the path as,
\begin{align}
\hat{D}(\mb{Z} \| \mb{X})
=
\hat{D}(\mb{Z} \| \mb{Z}^{k-1})
+
\cdots
+
\hat{D}(\mb{Z}^1 \| \mb{X}).
\end{align}
Accordingly, the respective (normalized) dual function, $\hat{f}_{\eta}(.)$, is also estimated indirectly as, 
\begin{align}
\hat{f}_{\eta}(\mb{x})
&=
\sum_{j=0}^{k-1}
f^{j}_{\eta}(\mb{x})
\\
\hat{f}^j_{\eta}(\mb{x})
&=
\hat{f}^j(\mb{x}) - \eta^f_j
\nonumber
\\
\eta^f_j
&=
\log
\sum_{\mb{z} \in \mb{Z}^{j}}
e^{f^{j}(\mb{z})}
- \log(|\mb{Z}^{j}|),
\nonumber
\label{eqn:dual_func_path}
\end{align}
    
with dual function $\hat{f}^j(.)$ corresponding from estimating $\hat{D}(\mb{Z}^{j+1} \| \mb{Z}^{j})$. In the following, we establish that sample complexity for estimating the dual function via the path can be exponentially smaller than from the direct estimate.
    
\paragraph{Sample complexity of the path-based estimator\\}
\hspace{-4.4mm} 
First, for the sake of completeness, we establish that the proposed divergence estimator via a path of dependency diffusion is consistent. See the supplement for more details.
% 
% While the above theorem holds for any base distribution $Q$, the following sample complexity result holds specifically for our choice of marginals as the base distribution.
% 
% However, for practical use of the estimator, the choice of base density matters as discussed next.
        % 
Despite the appeal of divergence estimation in its dual form, the reliability of these estimators has been a topic of debate in the field. One of the most important theoretical results in this regard is due to \cite{song2019understanding}, stating that the variance of the divergence estimate is lower bounded by the exponential of the true divergence.
% 
% This result applies to density estimation as well, and naturally raises concerns about practical applicability of the estimator if divergence is high w.r.t. base density. 
% 
% Since \kld of density P w.r.t. its product of marginals, Q, is lower than \kld w.r.t. Uniform density, our choice of base density is surely superior. However, estimating density directly from the product of marginals indeed has sample complexity lower bounded by exponential of the multi-variate mutual information in P, $\cI(P) \equiv \cD(P \| Q)$.  
    % 
In the following, we argue that it may be possible to \emph{avoid the exponential dependence} if we estimate it indirectly through the path of dependency diffusion as introduced in \secref{sec:dependency_diffusion_path}, in contrast to the direct estimate through a single step. 
\begin{algorithm}[tp!]
\caption{Samples via gradient walk in the dual space}
\begin{algorithmic}[1]
\REQUIRE{$\mb{X}$, $f(.)$}\\
\STATE $\mb{f} \gets$ f($\mb{X}$);\ \ $\mb{f} \gets$ sort($\mb{f}$) 
% \COMMENT{estimate the dual function for inputs}
% \STATE $\mb{f} \gets$ sort($\mb{f}$)
% 
% \COMMENT{sort inputs in the dual space}
% 
\STATE $\mb{d}_{knn} \gets$ estimateLocalDivergence($\mb{X}$)
\COMMENT{localized divergence estimation between kNNs in the dual space}
\STATE $\mb{c} \gets$ argpartition($\mb{d}_x$, c) 
\COMMENT{find indices of cut points with maximal local divergence between the neighbors}
% 
% \STATE $h_b(.) \gets$ histogramBins($\mb{f}$) 
% \COMMENT{histogram binning of }
\FOR{$j = 0 \to c$}
\STATE $\mb{X}^g_j \gets$ gradWalkDualSpace($f_{c_j}$, $f_{c_j+1}$) \COMMENT{generate samples between $f_{c_j}$ and $f_{c_j+1}$ per $\frac{df}{d\mb{x}}$}
\ENDFOR
\STATE \textbf{Return} $\mb{X}_s$ 
% \COMMENT{return generated samples}
\end{algorithmic}
\label{alg:gen_samples_gradient}
\end{algorithm}
\begin{theorem}
Variance for the direct estimation of ${D}(P \| Q)$  in its dual form using $n$ samples is,
\begin{align}
\lim_{n \to \infty}
n\ \mathrm{Var}_{P,Q} \hat{D}_n
&
\geq
-1
+
e^{D(P \| Q)}
\nonumber
\\
& 
=
-1
+
\prod_{j=0}^{d-2}
e^{D(Q^j \| Q^{j+1})},
\end{align}
whereas variance for its estimation via the dependency diffusion path, assuming the divergence estimates for each step to be independent, is:
\begin{align}
\lim_{n \to \infty}
n \mathrm{Var}_{P,Q}
\hat{D}_n^{ddp}
\geq
1-d +
\sum_{j=0}^{d-2}
e^{D(Q^j \| Q^{j+1})}
\label{eqn:variance_dependency_diffusion_path}
\end{align}
\end{theorem}
Here, in \eqnref{eqn:variance_dependency_diffusion_path}, the lower bound for the variance is no longer (exponentially) dependent upon the true measure of $D(P \| Q) \equiv I(P)$, rather dominated by the true measure of maximal divergence between two adjacent distributions on the path. Although it is not guaranteed that one inequality will strictly dominate the other in the worst-case scenario, one can assume that the overall dependency between all pixels in an image is usually greater than the dependency between two columns or rows within an image.

% In essence, by estimating divergence of real samples w.r.t. samples from the marginals via dependency diffusion, we avoid the otherwise exponential sample complexity which is especially prohibitive when there is high dependence (characterized by mutual information) between pixels~(or columns/rows) within images. 

% \subsubsection*{}
In reference to \figref{fig:dv_generative_sampling}, from estimating divergence of data distribution~(red dots) w.r.t. the marginal~(blue dots), we obtain global characterization (representation) of data distribution in the dual divergence space.
% 
% As we formally establish in \secref{sec:overall_algo_sampling}, any data point, with dual function value within the boundary of the red dots in the figure, is a valid in-distribution samples w.r.t. the data distribution.
% 
Next, we introduce the idea of localized divergence estimation between nearest neighbors in the dual space, so as to obtain a fine-grained characterization of the data distribution in the dual divergence space.
    
\subsection{Localized Divergence Estimation for Clustering}
\label{sec:local_div_est_knn}
    
Besides estimating divergence of the data distribution w.r.t. the respective marginal distribution, we propose to obtain a fine-grained characterization of the data distribution by learning clusters within the input dataset. In particular, we learn clusters such that divergence is maximized between the clusters as originally proposed by \cite{gargitc23}. This information-theoretic approach simplifies the problem of clustering to one of finding optimal cut points, the ones with maximal divergence, in the dual divergence space. See \figref{fig:localized_div_est_knn} and the pseudo-code in \algoref{alg:loc_div_est}. 

\begin{algorithm}[tp!]
\caption{Overall algorithm for generative sampling}
\begin{algorithmic}[1]
\REQUIRE{$\mb{X}$}
% 
% {Dataset of input (image) samples, $\mb{X} = \{ \mb{x}_i\}_{i=1}^n$, and hyperparameters, $t, t_w$}
% 
% \COMMENT{$t$ is number of iterations to learn the dual divergence estimator for sampling, and $t_w$ is the number of iterations for warming up the divergence estimator before starting to sample}\\
% 
\STATE $f(.) \gets$ initDualDivFunc($\mb{X}$)
\COMMENT{(neural) model of images as a dual function to estimate divergence}\\
\STATE $\mb{Z} \gets$ samplesFromMarginals($\mb{X}$)\\
% 
% \COMMENT{warmup divergence estimator to learn representation of real samples}
% 
\FOR{$i=0 \to t$}
\algblueshade{
% for characterizing the dependency structure between pixels}
% 
% \STATE $\mb{f}_x \gets$ f($\mb{X}$) \COMMENT{estimate the dual function for the real samples $\mb{X}$}
% \STATE $\mb{f}_z \gets$ f($\mb{Z}$) \COMMENT{estimate the dual function for samples $\mb{Z}$ from the marginals}
% \STATE $d_{xz} \gets$ D($\mb{f}_x$, $\mb{f}_z$) \COMMENT{estimate empirical divergence between $\mb{X}$ and $\mb{Z}$}
% 
\COMMENT{estimate divergence of inputs w.r.t. the marginals}
\STATE $\mb{f}_x \gets$ f($\mb{X}$),\ \ $\mb{f}_z \gets$ f($\mb{Z}$) 
% \COMMENT{estimate the dual function for $\mb{X}$}
% \STATE $\mb{f}_z \gets$ f($\mb{Z}$) 
% \COMMENT{estimate the dual function for $\mb{Z}$}
\STATE $d_{xz} \gets$ D($\mb{f}_x$, $\mb{f}_z$) \COMMENT{divergence between $\mb{X}$ and $\mb{Z}$}
% 
% \STATE $d_{xz} \gets$ divergenceToMarginals($\mb{X}$, $\mb{Z}$)
% \COMMENT{estimate empirical divergence of inputs w.r.t. samples from the marginals}
% 
\STATE $f(.) \gets$ backPropagate($f(.)$, $-d_{xz}$)
% \COMMENT{backpropagate to increase the divergence estimate}
}
\algredshade{
    % \STATE $\mb{f} \gets$ f($\mb{X}$) \COMMENT{estimate the dual function for the real samples $\mb{X}$}
    % \STATE $\mb{f} \gets$ sort($\mb{f}$)
    % \COMMENT{sort real samples in the dual divergence space}
    % \STATE $\mb{d}_x[j] \gets$ D($\mb{f}[j-k:j]$, $\mb{f}[j:j+k]$) $\forall j \in \{1, \cdots, n\}$
    % \COMMENT{localized divergence estimation between kNNs from the left and right hand side of each input in the dual divergence space.}
    % \STATE $l_c \gets \sum_j \mb{d}_x[j] - \log \sum_j e^{\mb{d}_x[j]}$ \COMMENT{multi-scale clustering loss }
    % 
    \COMMENT{local divergence estimation between neighbors}
    \STATE $\mb{d}_{knn} \gets$ estimateLocalDivergence($\mb{X}$)
    % \COMMENT{localized divergence estimation between kNNs in the dual space} 
    % \COMMENT{local divergence estimation between kNNs} 
    \STATE $l_c \gets \sum_j \mb{d}_{knn}[j] - \log \sum_j e^{\mb{d}_{knn}[j]}$
    \COMMENT{clustering loss from the local divergence estimation}
    \STATE $f(.) \gets$ backPropagate($f(.)$, $l_c$)
    % \COMMENT{backpropagate to reduce the clustering loss.}
}
\IF{$i \geq t_w$}
\algshade{
    \COMMENT{sample via gradient walk in the dual space}
    \STATE $\mb{X}^g \gets$ sampleViaGradientWalk($\mb{X}$, $f(.)$)
    % 
    % \COMMENT{generate samples in the dual divergence space}
    % 
    % % \STATE $\mb{f} \gets$ f($\mb{X}$);\ $\mb{f} \gets$ sort($\mb{f}$) \COMMENT{estimate dual function for real samples $\mb{X}$ and then sort the same.}
    % \STATE $\mb{d}_{knn} \gets$ localDivergenceEstimationkNN($\mb{X}$)
    % \COMMENT{localized divergence estimation between kNNs as above.}
    % \STATE $\mb{c} \gets$ argpartition($\mb{d}_x$, c) 
    % \COMMENT{find index of $c$ cut points with maximal local divergence}
    % \FOR{$j \in \mb{c}$}
    % \STATE $\mb{X}_s[j] \gets$ generateSamplesViaDualFuncGradient($\mb{f}[j]$, $\mb{f}[j+1]$)
    % \COMMENT{generate samples in the dual functional space between $\mb{f}[j]$ and $\mb{f}[j+1]$ as per gradient of the dual function w.r.t. input space}
    % \ENDFOR
    % 
    \STATE $\mb{f}_g \gets$ f($\mb{X}^g$) 
    % \COMMENT{estimate the dual function for generated samples}
    \STATE $\hat{D}_{xg} \gets$ D($\mb{f}_x$, $\mb{f}_g$) \COMMENT{divergence w.r.t. samples}
    \STATE $f(.) \gets$ backPropagate($f(.)$, $-\hat{D}_{xg}$)
    % 
    % \COMMENT{backpropagate for increasing the divergence estimate}
}
\ENDIF
\ENDFOR
\STATE $\mb{X}^g \gets$ sampleViaGradientWalk($\mb{X}$, $f(.)$)
\STATE \textbf{Return} $\mb{X}^g$ \COMMENT{return generated samples}
\end{algorithmic}
\label{alg:lrn_estimator_gen_samples_overall}
\end{algorithm}
        
As we show in \figref{fig:localized_div_est_knn}, we propose to have a more localized estimate of divergence at cut points, i.e. divergence between k-nearest neighbors on either side of the cut point. This approach is particularly well-suited to our problem context, where clustering is a secondary objective aimed at learning refined representations of data points in the dual divergence space. Moreover, for the same reason, the number of clusters $c$ in our problem setting is not a fixed (small) number. Rather, owing to the localized divergence estimation at cut points, we learn as many clusters as possible by maximizing softmax~($logsumexp$) of divergence on all the cut points while minimizing the mean statistic for the same. The former can be thought of as inter-cluster divergence and the latter as intra-cluster divergence.
Effectively, as illustrated in the figure, we obtain clustering at different scales. 
As shown in the figure, we can generate novel samples in empty dual space between the clusters.
    
% We refer to \cite{gargitc23} for more details.
% 
% As we discuss next in the overall algorithm, having characterized the data distribution locally as well as globally as per the ideas introduced above, we generate novel samples in the dual divergence as also shown in the figure above; see the pseudo code for sampling in \algoref{alg:gen_samples_gradient}.
    
% \begin{theorem}[nice to have but lower priority]
% \todo{extend clustering guaranties from uai 23 paper for the clustering objective here.}
% \end{theorem}
    
% Next, we discuss in details our overall algorithm for generating novel samples in the dual space.
        
\subsection{Overall Algorithm for Generative Sampling}
\label{sec:overall_algo_sampling}

In the previous sections, we have introduced two fundamental concepts for characterizing the data distribution in the dual space. The first concept involves estimating the divergence between the data distribution and its respective marginal distributions, in order to represent the dependencies between input dimensions (pixels). The second involves estimating the divergence locally between nearest neighbors, in order to achieve a fine-grained representation of data points within the data distribution.
Overall, in our proposed approach, we employ a (single) neural model that learns to estimate the divergences in an iterative manner. 

In \algoref{alg:lrn_estimator_gen_samples_overall}, in each iteration, the neural model updates its weights to increase the empirical estimate (in the dual form) of divergence of data distribution w.r.t. the marginals, as well as to decrease the clustering loss estimated from the localized divergence estimation. Further, within the same iteration, novel samples are generated in the presently optimized dual space via gradient walk in the empty spaces between the clusters (see \algoref{alg:gen_samples_gradient} for more details).
Note that the neural model learns to estimate the divergence of data distribution w.r.t. generated samples as well. The intuition behind the latter step is that some of the generated samples may be out of distribution w.r.t. the data distribution. This could arrive due to noisy gradients of the dual function w.r.t. inputs. Estimating divergence w.r.t. samples filters out the %allows rejection of such 
\gls{OOD} samples as well as robustifies the estimation of dual function gradient in the empty space between clusters.
% ~(in spirit to adversarial learning).
    
% In this manner, the divergence estimator is 
    
% \todo{provide more details on samples generation from theory perspective, describing how histograms are used to decide how many samples to generate between two clusters, and to provide intuitions that, as such, theoretical guarranties on divergence of generated samples w.r.t. data distribution is bounded by maximal gap between two observed data points.}
    
% \todo{Overall Algorithm Pseudocode for Generative Sampling}

% \subsubsection{A Gradient Walk between Cut Points}

% \subsubsection{Divergence w.r.t. Generated Samples}
% Local and global

We establish that the divergence of generated samples w.r.t. data distribution can be bounded as below.
\begin{theorem} 
An empirical estimate of the KL divergence between real samples $X$ and generated samples $X_g$ using~\eqnref{eq:dual_div_eq}, can be bounded as 
$\hat{D}(X_g \| X) = O(d^{knn}_{max})$ where $d^{knn}_{max}$ is the maximal distance of a point from $X_g$ to its nearest point in $X$ in the one-dimensional dual functional space of both the
sets.
\label{thm:div_gen_samples_wrt_input_samples}
\end{theorem}
% 
% \begin{proof}
% % 
%     Proof simply comes as an implication of Theorem 4 in \cite{garg2023or}. 
%     We have, 
%     \begin{align*}
%         \hat{D}(X\|X_g) = \max_{\hat{f} \in \mathcal{H}} \sum_{x_j \in X} \frac{f(x_j)}{M} - \log \sum_{\hat{x}_j \in X_g} \frac{e^{\hat{f}(\hat{x}^j)}}{N}.
%     \end{align*}
%   From the statement, we know that $d^{knn}_{max}$ is the maximal distance of a point from $X_g$ to its nearest point in $X$. Hence, if we create bins of size $d^{knn}_{max}$ in one dimension to put $f(x)$ for $x \in X$ and $X_g$, then everyone that contains $f(x)$ for $x \in X$ for any $x$ also contain an $f(\hat{x})$ for $\hat{x} \in X_g$. Hence, we do not need to discard any points from $X_g$ as in Theorem 4 in \cite{garg2023or}. Hence, we get $\hat{D}(X\|X_g) = O(d^{knn}_{max})$.
% \end{proof}
    % 
Since we generate novel samples in empty space between clusters in the dual space, as per the above theorem, it reduces the empirical divergence of generated samples w.r.t. real samples as a consequence of the reduction in the maximal \gls{knn} distance~($d^{knn}_{max}$) between the two sets.

\section{Experiments}

We conduct experiments for the problem of generating novel samples of  entire multivariate timeseries as images, using several datasets from a diverse set of domains as described below.

\begin{figure*}
\centering
\begin{subfigure}{0.8\textwidth}
\includegraphics[width=.99\linewidth]{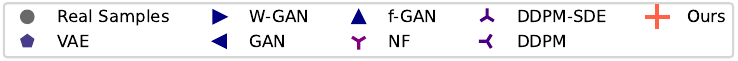}
\end{subfigure}
\begin{subfigure}{.25\textwidth}
\includegraphics[width=.99\linewidth]{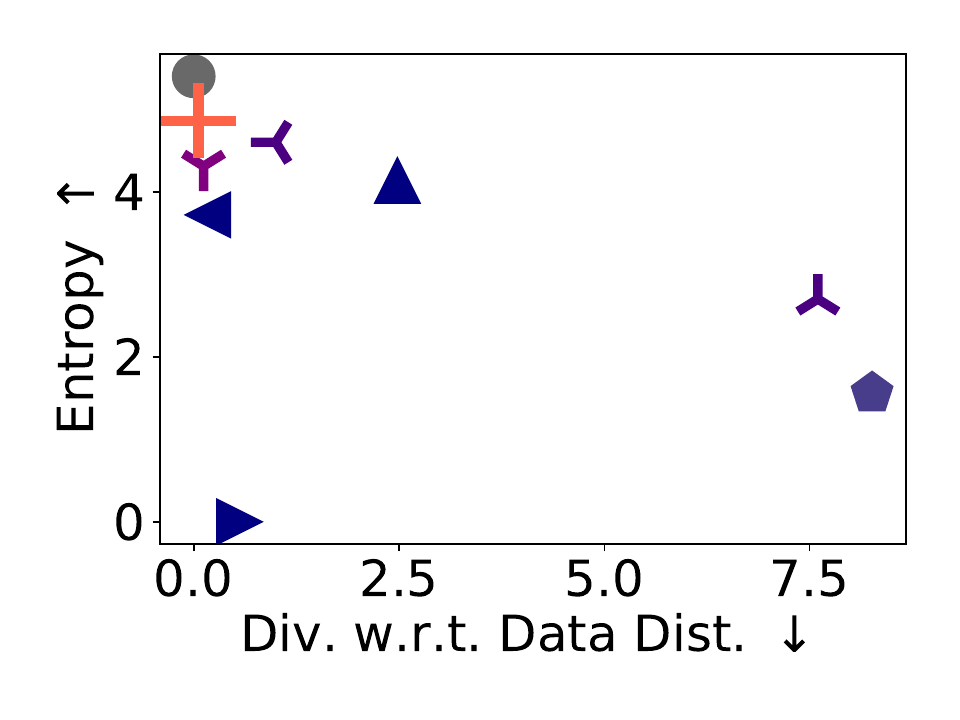}
\caption{EEG}
\end{subfigure}
\hspace{-3mm}
\begin{subfigure}{.25\textwidth}
\includegraphics[width=.99\linewidth]{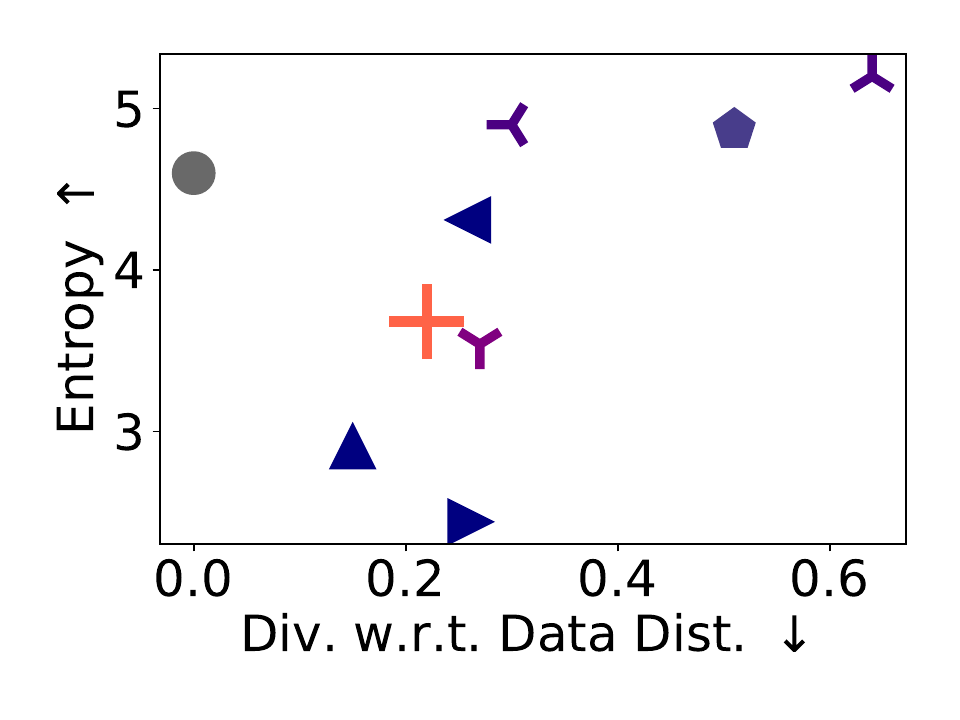}
\caption{Neuropixels}
\end{subfigure}
\hspace{-3mm}
\begin{subfigure}{.25\textwidth}
\includegraphics[width=.99\linewidth]{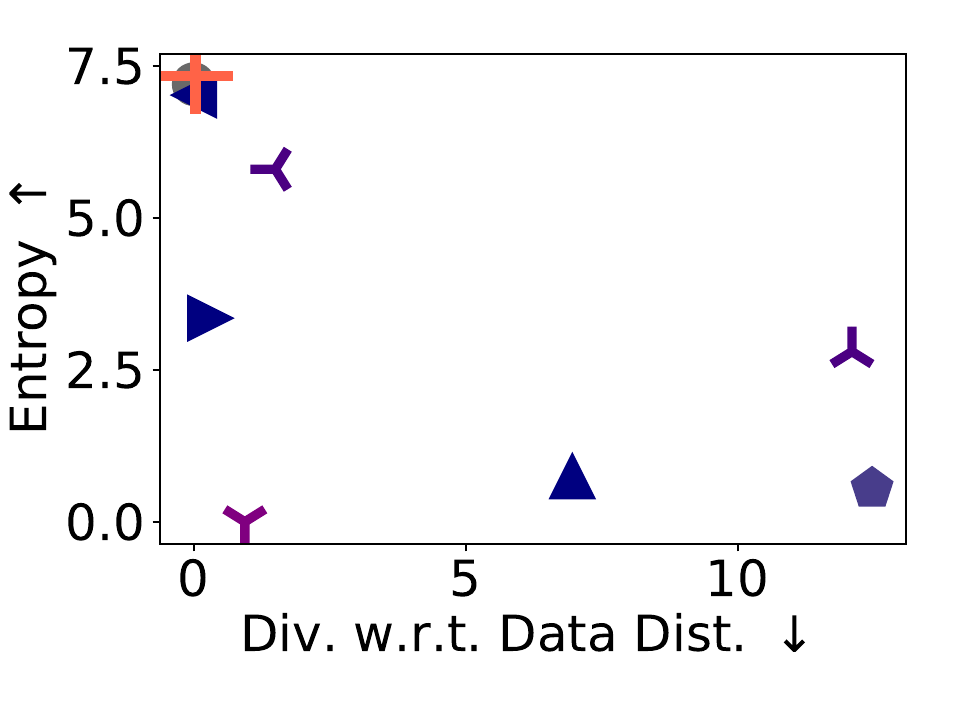}
\caption{Stock Returns}
\end{subfigure}
\hspace{-3mm}
\begin{subfigure}{.25\textwidth}
\includegraphics[width=.99\linewidth]{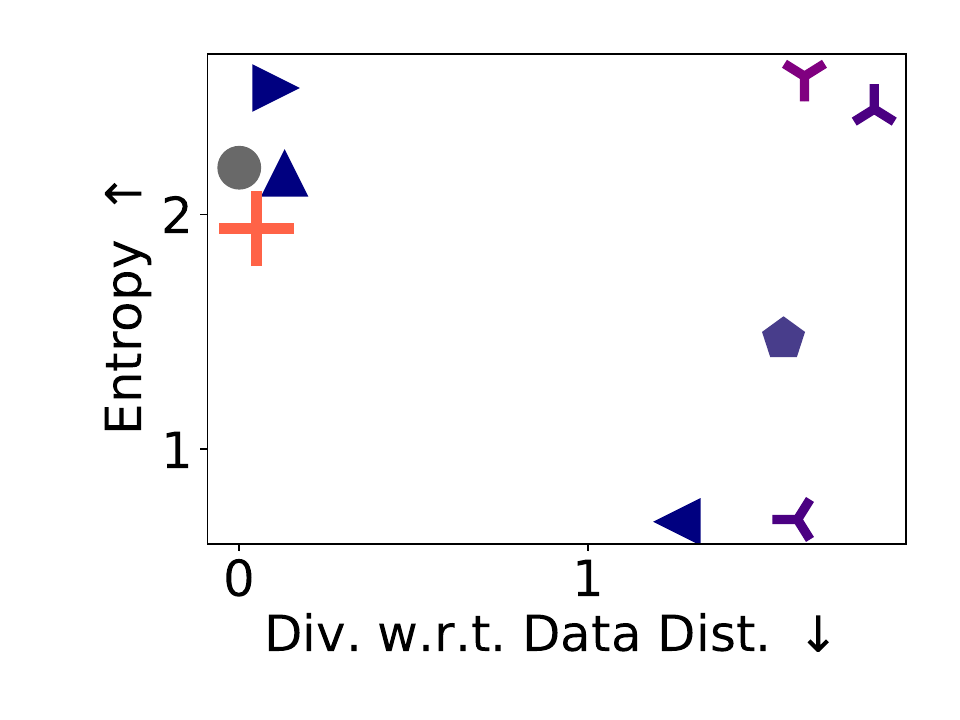}
\caption{City Traffic}
\end{subfigure}
\begin{subfigure}{.25\textwidth}
\includegraphics[width=.99\linewidth]{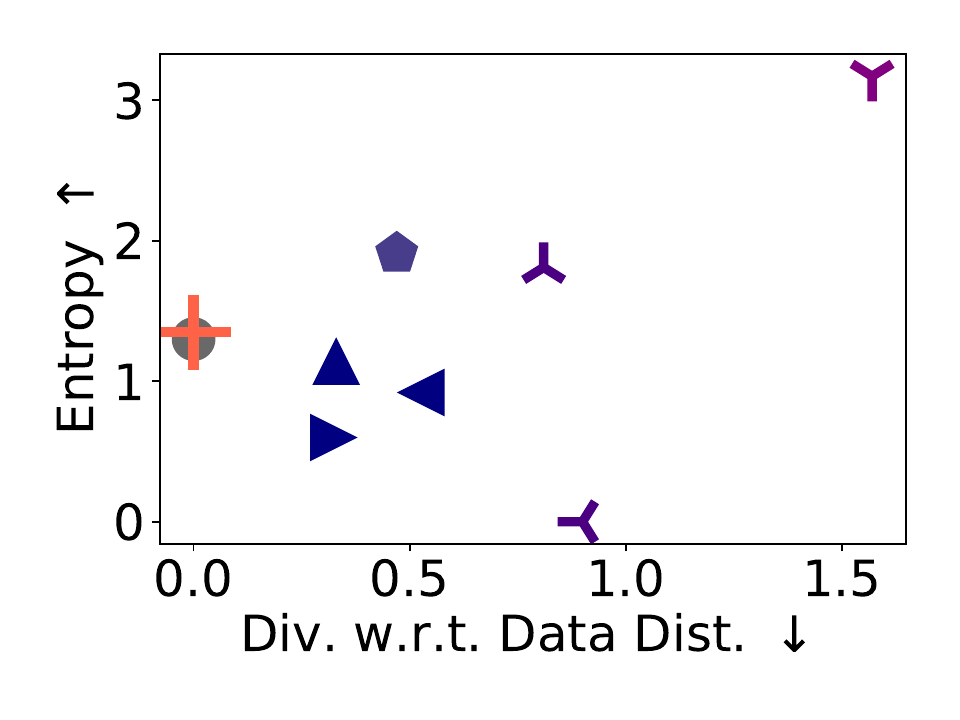}
\caption{Air Pollution}
\end{subfigure}
\hspace{-3mm}
\begin{subfigure}{.25\textwidth}
\includegraphics[width=.99\linewidth]{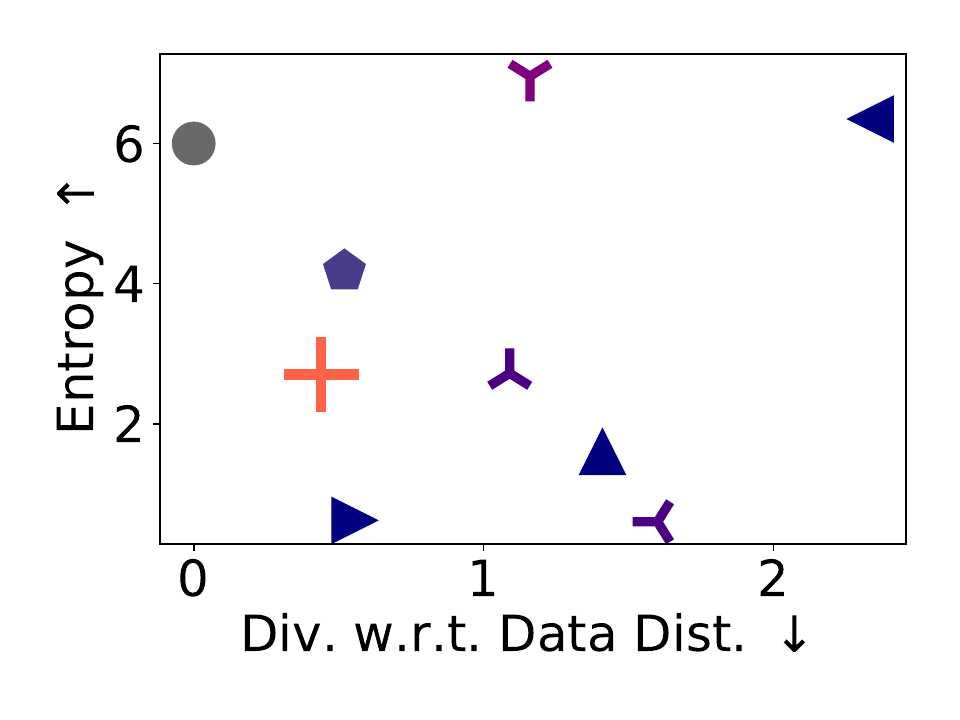}
\caption{Wind Farms}
\end{subfigure}
\hspace{-3mm}
\begin{subfigure}{.25\textwidth}
\includegraphics[width=.99\linewidth]{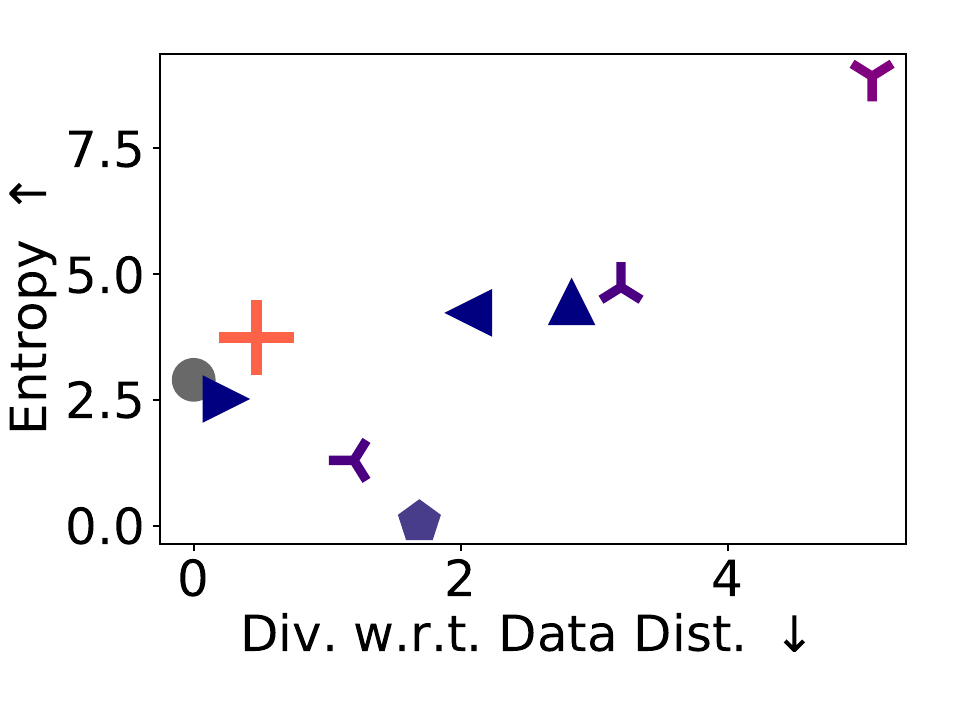}
\caption{Solar Energy}
\end{subfigure}
\hspace{-3mm}
% 
% \begin{subfigure}{.33\textwidth}
% \includegraphics[width=.99\linewidth]{gen_samples_eval/Precipitation.pdf}
% \caption{Precipitation}
% \end{subfigure}
% % 
% \hspace{-3mm}
% 
\begin{subfigure}{.25\textwidth}
\includegraphics[width=.99\linewidth]{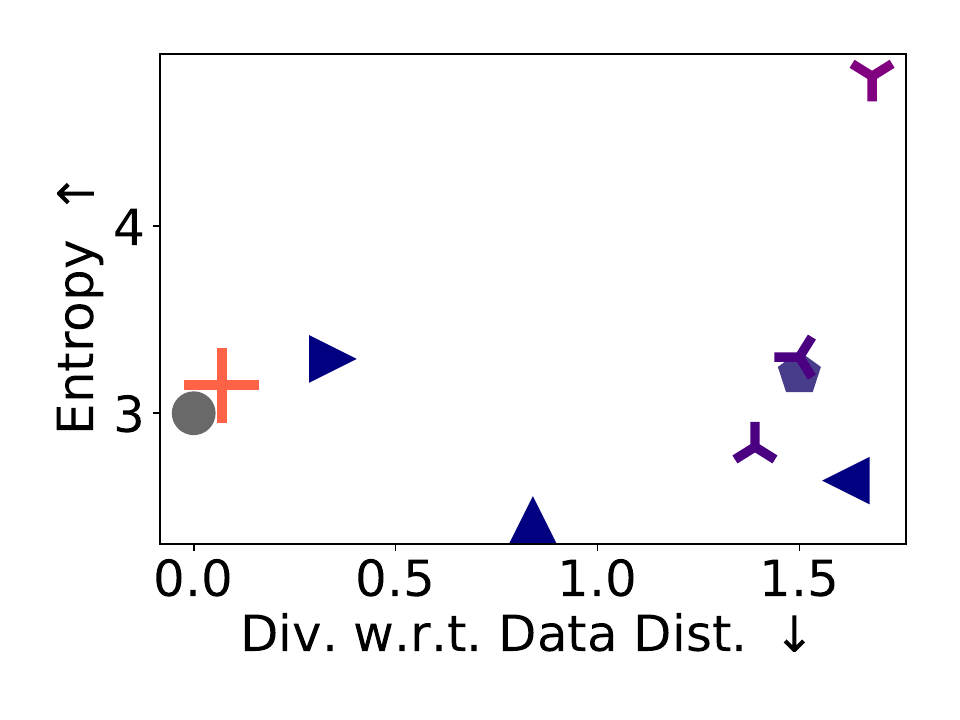}
\caption{Electrical Consumption}
\end{subfigure}
% 
% \hspace{-3mm}
% % 
% \begin{subfigure}{.33\textwidth}
% \includegraphics[width=.99\linewidth]{gen_samples_eval/IndoorTemperature.pdf}
% \caption{Indoor Temperature}
% \end{subfigure}
% 
\caption{All the methods are compared in terms of KL-Divergence of generated samples w.r.t. data distribution which is desired to be minimized while maximizing entropy of the generated samples.}
\label{fig:generative_sampling_entropy_div_wrt_data_dist}
\end{figure*}

\paragraph{Datasets}
% 
% \hspace{-4.4mm}
% 
(i) \textbf{EEG}: We are interested in analyzing the electrical activity in the brains of 81 patients diagnosed with schizophrenia, utilizing data from 74 channels per patient. %human brains from 74 channels for each of the 81 patients diagnosed with schizophrenia. Correspondingly, 
After preprocessing, we obtain 81 real samples of images each of dimension 72x72.
(ii) \textbf{Neuropixels}: We model spiking neural activity over a 2-second period, capturing data from diverse regions within mouse brains using 443 high-density extracellular electrophysiology probes~\cite{siegle2021survey}. Derived from numerous recording sessions, we acquired 195 preprocessed images, each of dimension 96x96. The preprocessing included the exclusion of probes exhibiting minimal activity. %Corresponding from multiple trials of recordings, we obtained 195 images of size 96x96 after preprocessing including removal of probes with negligible activity.
(iii) \textbf{Stock Returns:} We analyze returns for the 50 most liquid securities from the Russell 3000 index with each month from the past corresponding to a single image resulting in a dataset of 160 images, each of size 48x48. 
(iv) \textbf{City Traffic:} In the city of San Francisco, it is valuable to study hourly traffic across multiple sites in the city. Viewing weekly traffic %for an entire week 
as a single image, we obtain a dataset of 104 images, each of dimension 168x168.
(v) \textbf{Air Pollution:} Air quality indices are analyzed on an hourly basis for major cities across the world~(aka KDD Cup 2018). Similar to the traffic dataset, we view weekly pollution data as a single image and obtain a dataset of 56 images, each of dimension 168x168.
(vi) \textbf{Wind Farms:} Wind power production is recorded across wind farms in Australia. Analyzing it every 12 hours as an image, we obtain a dataset of 732 images, each of size 144x144.  
(vii) \textbf{Solar Energy:} Solar power data is measured every 10 minutes at many eastern U.S. locations. Considering the high inter- and intra-day dynamics of solar energy, we preprocess the data to analyze solar energy variation in a day as a single image. This results in a dataset with 386 images, each of size 136x136. %after preprocessing. 
(viii) \textbf{Electrical Consumption:} Electricity consumption was recorded every hour at 370 sites from year 2012 to 2014. We preprocess the data and obtain the weekly electricity consumption as a single image. This results in a dataset of 156 images, each of size 168x168.
% 
% (ix) \textbf{Berkeley Indoor Temperature:} Indoor temperature was recorded via 46 wireless sensors in Intel laboratory at Berkeley. After preprocessing, we obtain 180 images of size 40x40. 
% 
(Additional information about these datasets is available in the supplement.)
%More details on these datasets can be found in the supplement. 
% 
% As it is a standard practice, all training images are normalized before learning a generative method on the same.
    
% used public electrophysiological Neuropixels dataset~\cite{siegle2021survey}. 
% 
% [Siegle et al., 2021, Institute, 2020]. Multiple high-density extracellular electrophysiology probes were used to simultaneously record spiking neural activity from a wide variety of areas in the mouse brain. We used the data of the animal with session-id 798911424 and included the first 100 out of 195 trials. The first 2000 ms of each trial after stimulus onset was extracted. We time-binned the timestamps with 0.1 ms resolution, giving 443 timeseries, each of length 20,000 timesteps.
    
% \begin{figure*}[ht!]
% \centering
% % \includegraphics[width=1.9\columnwidth,scale=0.2]{legend.png}
% \subfigure[Neuropixels]{
% \includegraphics[width=0.45\columnwidth,scale=0.3]{Figures/NeuroPixels_Dataset.pdf}
% \label{fig:neuropixels_mi}
% }
% % 
% \caption{Evaluating clusters in terms of pairwise mutual information between timeseries within clusters~(intra-cluster $\uparrow$) and across clusters~(inter-cluster $\downarrow$). The proposed method is ITC-DM shown in solid red circles.}
% \label{fig:expr_mi}
% \end{figure*}

\paragraph{Baselines}
To illustrate the efficacy of the proposed approach, we evaluate its performance against %We compare the proposed approach w.r.t. all 
various established methods for deep generative sampling of images. Specifically, we employ (i) VAE~\cite{kingma2013auto}, (ii) GAN~\cite{goodfellow2014generative}, (iii) WGAN~\cite{arjovsky2017wasserstein}, (iv) f-GAN~\cite{nowozin2016f}, (v) NF~\cite{papamakarios2017masked}, (vi) DDPM~\cite{ho2020denoising}, and (vii) DDPM via \gls{SDE}~(DDPM-SDE)~\cite{song2020score}.
    
\paragraph{Evaluation Settings}
For each method of generative sampling, we explore different choices of architectures for modeling including \gls{CNN}, \gls{ViT}, or even \gls{FNN} in some cases. In addition to standard encoding of patches within images, we also found it useful to encode all the rows and columns within an image~(aka horizontal or vertical patches). While leveraging the official implementations of the baseline methods, considering the (very) small size of training sets, we utilized different strategies to avoid overfitting including dropout, $l_2$ regularization, and lighter-weight architecture in terms of the number of layers, among others (see more details in the supplement). 
For each dataset, we generate 1000 samples from every method and evaluate the samples in terms of diversity, novelty, and being \gls{ID} w.r.t. data distribution as discussed next.

\begin{figure*}
\centering
\begin{subfigure}{0.8\textwidth}
\includegraphics[width=.99\linewidth]{gen_samples_eval/legend.pdf}
\end{subfigure}
\begin{subfigure}{.25\textwidth}
\includegraphics[width=.99\linewidth]{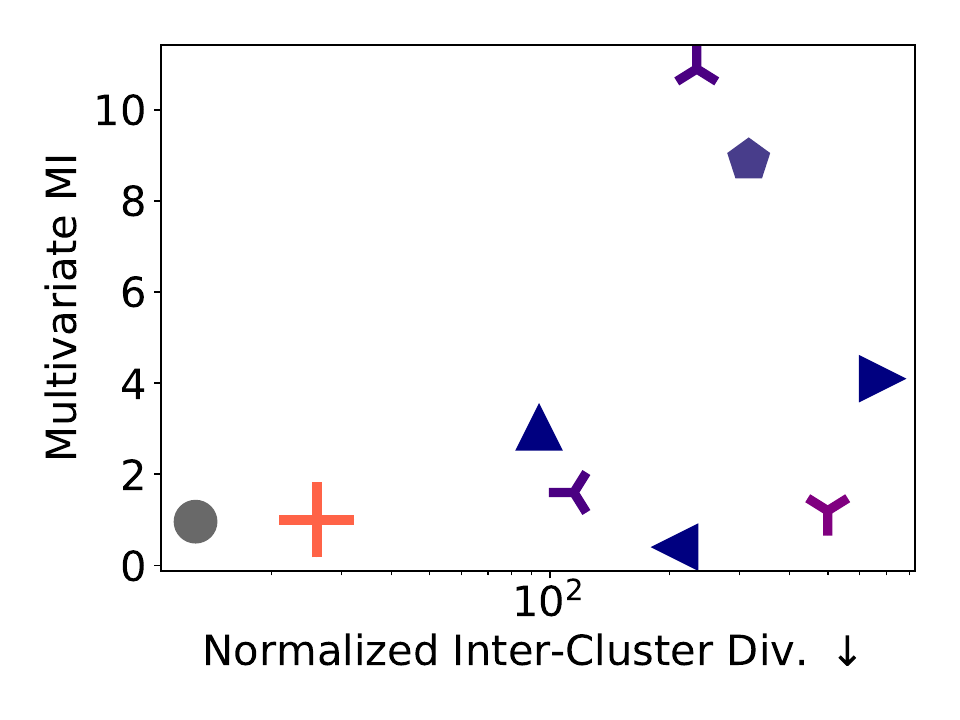}
\caption{EEG}
\end{subfigure}
\hspace{-3mm}
\begin{subfigure}{.25\textwidth}
\includegraphics[width=.99\linewidth]{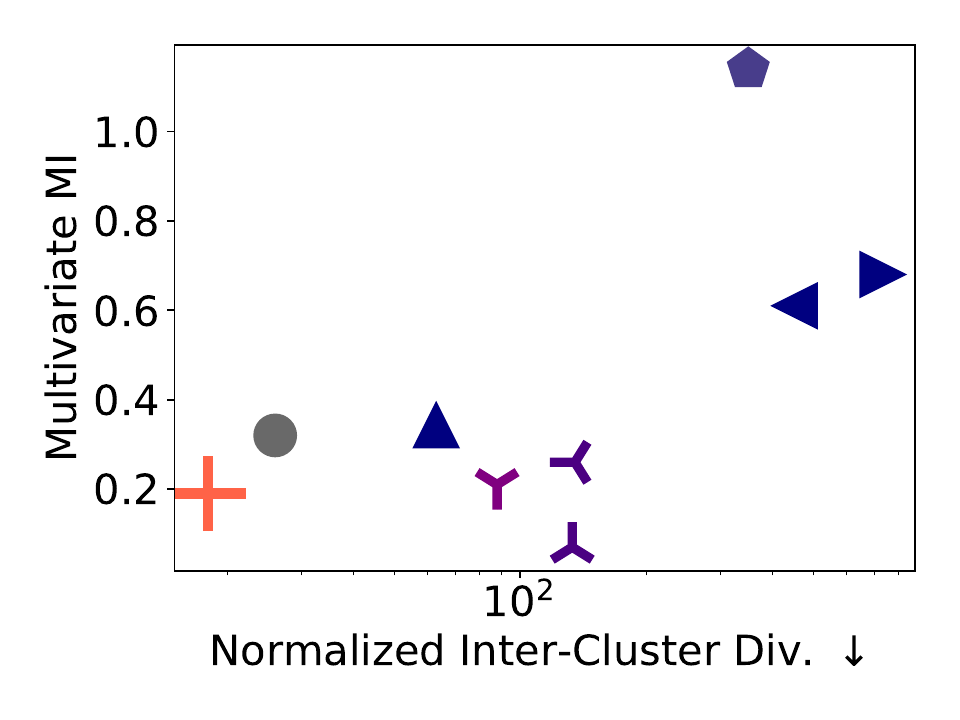}
\caption{Neuropixels}
\end{subfigure}
\hspace{-3mm}
\begin{subfigure}{.25\textwidth}
\includegraphics[width=.99\linewidth]{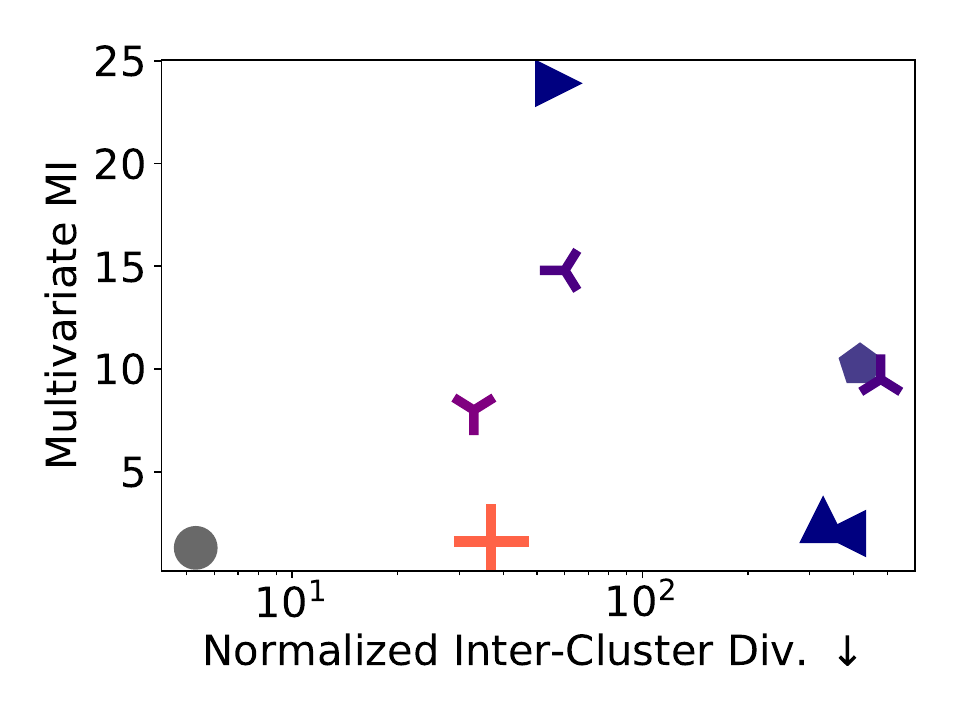}
\caption{Stock Returns}
\end{subfigure}
\hspace{-3mm}
\begin{subfigure}{.25\textwidth}
\includegraphics[width=.99\linewidth]{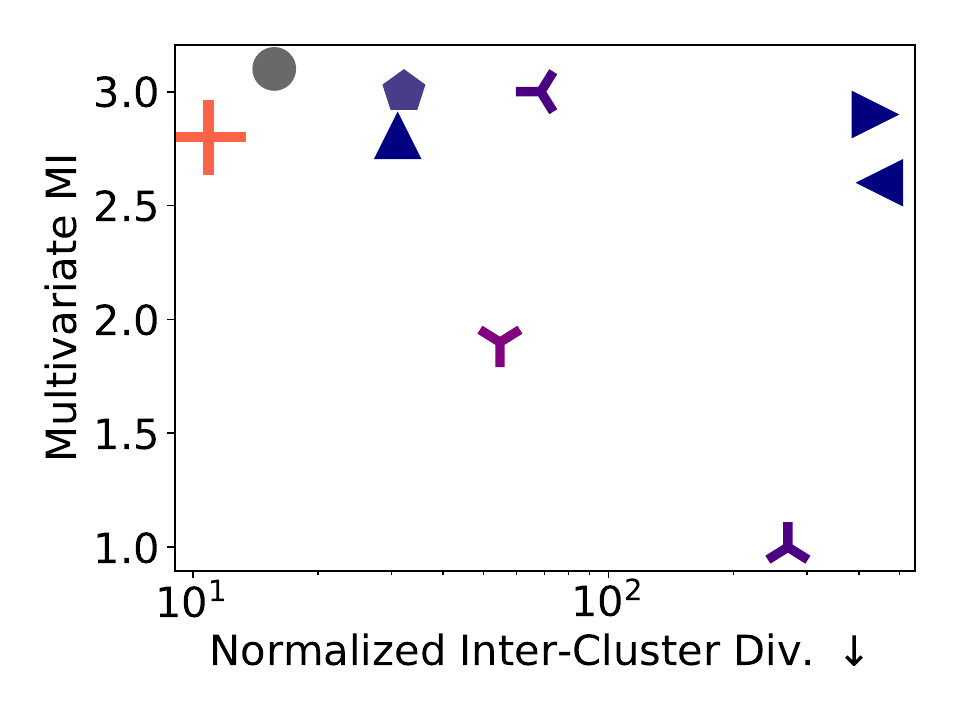}
\caption{Traffic}
\end{subfigure}
% 
% \hspace{-3mm}
% 
\begin{subfigure}{.25\textwidth}
\includegraphics[width=.99\linewidth]{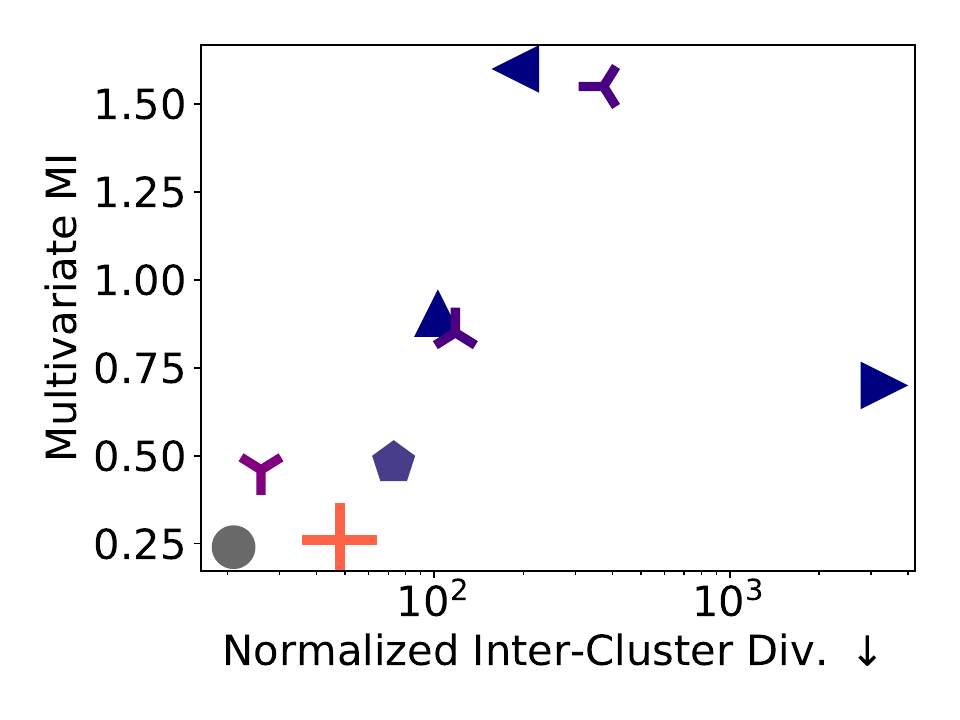}
\caption{Pollution}
\end{subfigure}
\hspace{-3mm}
\begin{subfigure}{.25\textwidth}
\includegraphics[width=.99\linewidth]{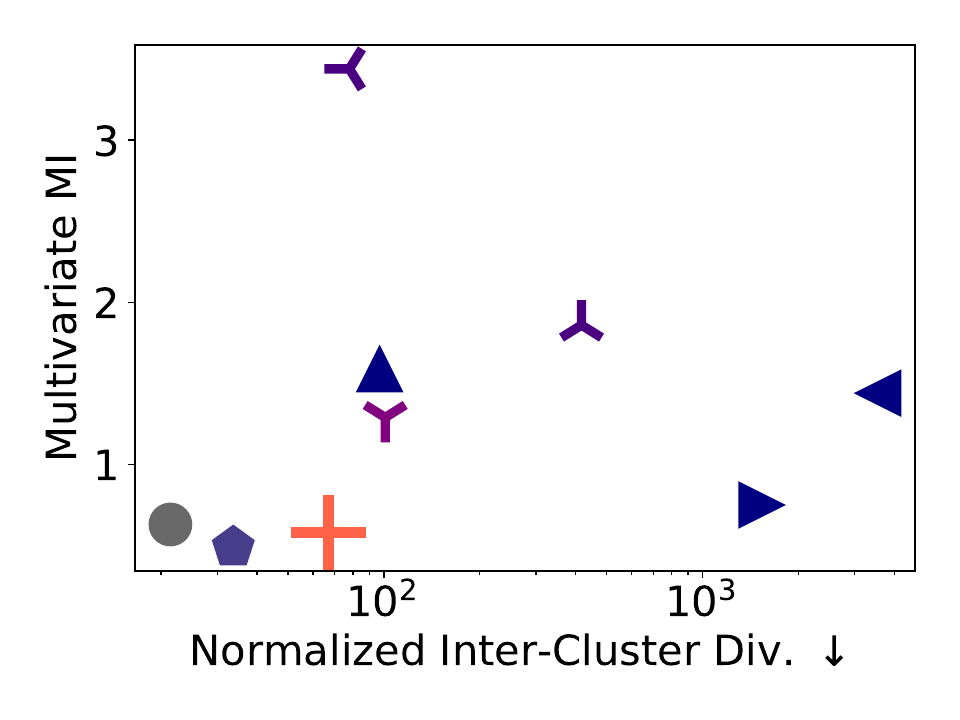}
\caption{Wind Farms}
\end{subfigure}
\hspace{-3mm}
\begin{subfigure}{.25\textwidth}
\includegraphics[width=.99\linewidth]{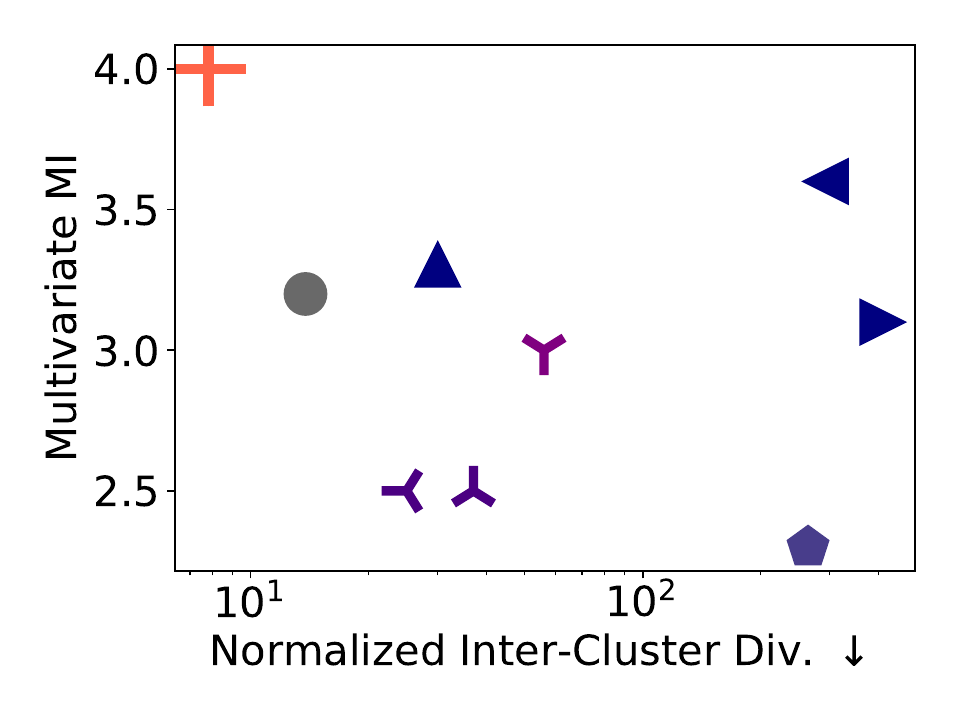}
\caption{Solar Energy}
\end{subfigure}
\hspace{-3mm}
% 
% \begin{subfigure}{.33\textwidth}
% \includegraphics[width=.99\linewidth]{gen_samples_eval/Precipitation.pdf}
% \caption{Precipitation}
% \end{subfigure}
% % 
% \hspace{-3mm}
% 
\begin{subfigure}{.25\textwidth}
\includegraphics[width=.99\linewidth]{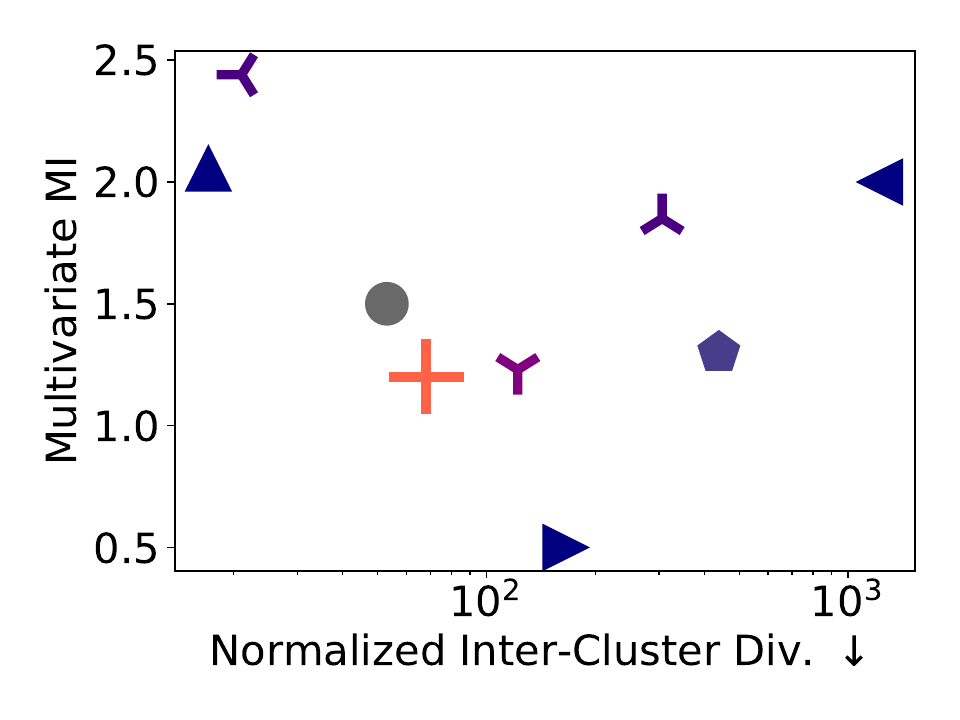}
\caption{Electrical Consumption}
\end{subfigure}
\caption{All the methods are evaluated based on two metrics. On the X-axis, we plot (normalized) maximal (empirical) divergence between clusters within the union set of input and generated samples. This measure is desired to be minimized~($\downarrow$), as an indirect proxy for the diversity of generated samples vs real samples. On the Y-axis, we plot estimated \gls{MMI} between pixels of images~(dimensions of samples). \gls{MMI} for generated samples should close in value to the \gls{MMI} for real samples.}
\label{fig:generative_sampling_mi_cluster_div}
\end{figure*}
    
\subsection{Evaluation Results}
    
%Evaluating methods for the problem of generating samples of images is as challenging as the problem itself, both being tied to each other.
    % 
Assessing methods for generating image samples is as intricate as the problem itself, given their inherent interdependence. Unlike the case of natural images where a pre-trained \gls{DNN}~(such as InceptionV3) is available for obtaining feature representations of generated images, modeled as multivariate Gaussian so as to measure \gls{FID} between the two distributions of images (real and generated)~\cite{heusel2017gans}, there is no such standardized model to obtain representation of images for our problem. Moreover, assuming the representations of images from a \gls{DNN} like InceptionV3 to be Gaussian distributed may suffice in practice for natural images, but not necessarily for other possible kinds of images like the ones considered in this paper. Nevertheless, we present a comparison of all the methods as per an \gls{FID} score in \tabref{tab:all_sampling_fid}, establishing superiority of our approach upon all the baselines.~\footnote{Here, the model for obtaining representation of images is the one which estimates mutual information between all the pixels in an image.} 

\begin{table*}[tp!]
\centering
\csize
\begin{tabular}{llllllllllll}
\toprule
Dataset&VAE&W-GAN&GAN&f-GAN&NF&DDPM-SDE&DDPM&\textbf{Ours}\\
\toprule
EEG&1.03$\pm$0.0&0.61$\pm$0.03&0.71$\pm$0.09&1.34$\pm$0.17&0.15$\pm$0.02&1.2$\pm$0.02&0.34$\pm$0.04&\textbf{0.12$\pm$0.06}\\
\midrule
NeuroPixels&6.44$\pm$0.00&0.78$\pm$0.02&1.85$\pm$0.05&1.0$\pm$0.04&0.97$\pm$0.03&3.71$\pm$0.10&1.19$\pm$0.10&\textbf{0.26$\pm$0.05}\\
\midrule
Stock Returns&5.21$\pm$0.02&4.43$\pm$0.09&1.2$\pm$0.08&2.16$\pm$0.25&10.43$\pm$0.30&5.78$\pm$0.10&3.8$\pm$0.15&\textbf{0.32$\pm$0.21}\\
\midrule
Traffic&4.46$\pm$0.11&3.49$\pm$0.50&26.42$\pm$0.98&1.65$\pm$0.26&9.85$\pm$0.58&36.93$\pm$0.10&20.16$\pm$0.18&\textbf{0.51$\pm$0.12}\\
\midrule
% KDD cup 2018
Pollution&0.22$\pm$0.04&0.11$\pm$0.0&0.25$\pm$0.01&0.34$\pm$0.02&0.73$\pm$0.02&0.78$\pm$0.01&0.15$\pm$0.01&\textbf{0.03$\pm$0.00}\\
\midrule
Wind Farms&1.31$\pm$0.06&6.85$\pm$0.12&19.29$\pm$0.0&10.35$\pm$0.34&4.11$\pm$0.07&12.2$\pm$0.04&10.56$\pm$0.33&\textbf{0.48$\pm$0.09}\\
\midrule
Solar Energy&25.88$\pm$0.0&0.85$\pm$0.07&4.58$\pm$0.60&9.4$\pm$0.21&42.91$\pm$0.59&26.74$\pm$0.03&9.55$\pm$0.28&\textbf{0.70$\pm$0.18}\\
\midrule
Electrical Consumption&4.94$\pm$0.02&1.1$\pm$0.06&9.56$\pm$0.01&23.15$\pm$0.27&8.2$\pm$0.40&18.06$\pm$0.05&4.8$\pm$0.25&\textbf{0.36$\pm$0.06}\\
\toprule
\end{tabular}
\caption{FID scores~($\downarrow$) for generated samples. For a set of generated samples from a method which is larger in size~(1000) than the original sample size, we randomly select samples from the set as many as the real samples to obtain FID score. Having performed 100 trials, we obtain mean and standard deviation of FID scores for a given method. Our method obtains the lowest FID scores w.r.t. all the other methods.
% 
% However, we advocate to not use this metric as the only measure to assess the quality of generated samples. For the same reason, we provided evaluation of all the methods with additional four metrics in the paper.
% 
}
\label{tab:all_sampling_fid}
\end{table*}
    
Because of the aforementioned limitations of FID metric, for the primary analysis, we introduce four advanced (information theoretic) evaluation metrics, estimated without any distribution assumptions or a pre-trained network, as discussed below.
The first metric measures the (empirical) divergence of generated samples w.r.t. data distribution~(i.e. training set of real samples). It aims to verify whether the generated samples accurately represent the underlying data distribution. %so as to validate if the generated samples represent the data distribution. This metric plays the same role as \gls{FID} but is estimated directly on raw images without any distributional assumptions.
While akin to \gls{FID}, this metric is computed directly on raw images without making any distributional assumptions. %However, this metric can be misleading on its own for the scenarios of overfitting. 
Nevertheless, relying solely on this metric can be misleading in cases of overfitting. To address this, we propose to measure the (empirical) entropy of generated samples, estimated via its KL-divergence w.r.t. uniform distribution, as a proxy for the diversity of samples. Before introducing the other two metrics, in reference to \figref{fig:generative_sampling_entropy_div_wrt_data_dist}, we first discuss our evaluation results as per these two metrics. For datasets, \gls{EEG}, stock returns, and electricity consumption, our approach dominates, exhibiting low divergence and high entropy. In the context of the traffic dataset, our method proves to be highly competitive, with W-GAN and f-GAN generating samples of marginally higher entropy (diversity). However, for the wind farms dataset, GAN outperforms our method, and NF is also highly competitive, albeit with a trade-off of %though at the expense of 
higher divergence of generated samples w.r.t. data distribution. In the case of the solar energy dataset, our approach and W-GAN demonstrate equal competitiveness. DDPM stands out as competitive only for the EEG, neuropixels, and stock returns datasets.
    
To measure the novelty of generated samples w.r.t. real samples, we introduce a metric in relation to \thmref{thm:div_gen_samples_wrt_input_samples}, i.e. (normalized) softmax of divergence between clusters within the joint set of real samples and generated samples. The metric value is inversely proportional to the novelty of the generated samples w.r.t. real samples. In addition, we are also interested in analyzing if generated (image) samples characterize dependencies between pixels as in the original real samples. While it is not possible to estimate dependencies explicitly for comparison between the two sets, we quantify \gls{MMI} between pixels for real samples vs generated samples. We expect generated samples to have the same \gls{MMI} value as the real samples. See the comparison between methods as per these two metrics in \figref{fig:generative_sampling_mi_cluster_div}. For datasets, EEG, neuropixels, stock returns, and traffic, our method generates samples with higher novelty w.r.t. other methods while maintaining \gls{MMI} similar to real samples. For the pollution dataset, when comparing NF to our method, there is a trade-off. A similar trade-off appears for electrical consumption and solar energy datasets. For the wind farms dataset, our method is the most competitive after GAN.
    
% \paragraph{Visualization of images}
% 
    
\paragraph{Evaluation using supplementary metrics}
Besides the primary evaluation as discussed above, in \tabref{tab:supplement_eval_metrics}, we present our analysis using four supplementary evaluation metrics. The first metric is Inception score (aka IS in the literature) as a proxy for the diversity of samples. This is estimated using the original Inception-v3 model itself. Using Inception-v3, we also estimate the most popular metric FID. Note that the FID scores reported above in \tabref{tab:all_sampling_fid} are different from the ones presented here in \tabref{tab:supplement_eval_metrics} as the former ones are estimated using a neural MI estimator.
Moreover, similar in spirit to the KL-divergence metric (\figref{fig:generative_sampling_entropy_div_wrt_data_dist}), we evaluate generated samples in terms of the Wasserstein distance aka WD~(as in W-GAN) w.r.t. the real samples. Lastly, we estimate negative log likelihood (NLL) of generated samples using a DDPM. 
Overall, across all the metrics and 8 datasets, our method consistently outperforms the competitive approaches. 

\begin{table}[tp!]
\centering
\csize
\renewcommand{\tabcolsep}{1.6pt}
\begin{tabular}{llllllllllll}
\toprule
Data&Metric&VAE&wGAN&GAN&fGAN&NF&DDPMsde&DDPM&\textbf{Ours}\\
\toprule
EEG&IS$\uparrow$&1.0&1.3&\textit{2.6}&5.4&2.0&1.5&1.9&\textbf{3.1}\\
&FID$\downarrow$&0.52&0.28&\textit{0.13}&1.04&0.15&0.40&0.25&\textbf{0.11}\\
&WD$\downarrow$&8.45&3.29&0.27&2.65&\textbf{0.13}&14.41&0.50&\textit{0.26}\\
&NLL$\downarrow$&0.18&0.05&0.05&0.46&0.06&0.21&\textit{0.03}&\textbf{0.01}\\
\midrule
NP&IS$\uparrow$&1.2&2.1&3.2&2.5&2.5&2.3&\textbf{3.8}&\textit{2.8}\\
&FID$\downarrow$&0.56&0.40&0.69&0.46&0.29&0.49&\textit{0.14}&\textbf{0.10}\\
&WD$\downarrow$&0.87&0.28&0.42&0.18&\textit{0.12}&0.15&0.16&\textbf{0.05}\\
&NLL$\downarrow$&0.24&\textbf{0.05}&0.15&0.12&0.13&0.24&\textit{0.06}&0.07\\
\midrule
SR&IS$\uparrow$&1.0&1.4&1.1&1.2&1.5&1.5&\textit{1.4}&\textbf{3.4}\\
&FID$\downarrow$&1.22&0.24&0.19&1.34&0.37&1.11&\textit{0.14}&\textbf{0.05}\\ 
&WD$\downarrow$&12.62&20.17&\textbf{0.43}&7.23&48.94&30.86&8.53&\textit{3.32}\\
&NLL$\downarrow$&0.15&0.08&\textbf{0.01}&0.41&0.09&0.19&\textbf{0.01}&\textit{0.03}\\
\midrule
CT&IS$\uparrow$&2.2&1.6&2.2&2.2&\textit{3.4}&1.8&2.3&\textbf{4.7}\\
&FID$\downarrow$&0.53&0.31&0.88&0.36&\textit{0.23}&1.14&0.33&\textbf{0.08}\\
&WD$\downarrow$&1.57&\textbf{0.10}&1.29&\textit{0.13}&0.74&1.43&1.67&0.15\\
&NLL$\downarrow$&0.12&0.08&0.09&\textit{0.06}&0.14&0.34&\textit{0.06}&\textbf{0.01}\\
\midrule
% KDD cup 2018
% 
AP&IS$\uparrow$&\textit{10.4}&1.7&1.7&4.5&4.5&1.8&\textit{10.4}&\textbf{10.6}\\
&FID$\downarrow$&0.74&1.07&1.12&0.65&0.77&1.04&\textit{0.53}&\textbf{0.15}\\
&WD$\downarrow$&0.58&0.43&1.11&0.75&\textit{0.34}&0.86&1.33&\textbf{0.02}\\
&NLL$\downarrow$&0.05&0.06&0.12&0.19&0.21&0.29&\textit{0.03}&\textbf{0.01}\\
\midrule
WF&IS$\uparrow$&4.9&3.0&1.0&2.5&\textbf{6.4}&2.3&\textit{5.9}&\textit{5.9}\\
&FID$\downarrow$&0.57&0.78&1.32&1.55&0.63&1.13&\textit{0.48}&\textbf{0.17}\\
&WD$\downarrow$&\textit{0.28}&0.66&1.15&1.51&1.26&1.21&1.53&\textbf{0.23}\\
&NLL$\downarrow$&0.04&0.05&0.29&0.33&0.07&0.30&\textit{0.02}&\textbf{0.01}\\
\midrule
SE&IS$\uparrow$&1.0&\textit{4.3}&2.9&3.5&3.4&1.7&\textbf{10.6}&\textit{5.8}\\
&FID$\downarrow$&1.77&\textit{0.21}&\textbf{0.16}&1.4&0.75&1.34&0.32&0.25\\
&WD$\downarrow$&1.87&\textbf{0.37}&1.96&2.88&4.60&3.23&\textit{1.39}&1.71\\
&NLL$\downarrow$&\textbf{0.01}&\textit{0.02}&0.03&0.42&0.18&0.32&\textit{0.02}&\textbf{0.01}\\
\midrule
EC&IS$\uparrow$&1.2&1.6&1.0&2.0&\textit{3.1}&1.8&\textbf{3.8}&2.0\\
&FID$\downarrow$&1.45&\textit{0.88}&1.52&1.29&1.11&1.63&1.02&\textbf{0.05}\\
&WD$\downarrow$&0.54&\textit{0.25}&1.65&1.82&0.61&1.72&1.94&\textbf{0.10}\\
&NLL$\downarrow$&0.25&0.07&0.19&0.42&0.14&0.29&\textit{0.04}&\textbf{0.01}\\
\toprule
\end{tabular}
% 
% \vspace{-3.0mm}
% 
\caption{Analysis using supplementary evaluation metrics. Due to space constraint, dataset names are abbreviated. 
% 
% (NP for Neuropixels, SR for Stock Returns, CT for City Traffic, AP for Air Pollution (aka KDD Cup 2018), WF for Wind Farms, SE for Solar Energy, and EC for Electrical Consumption).
% 
The best numbers (within rows) are shown in bold and the second best are in italic. 
% 
% (FID, Inception) scores~($\downarrow$) computed as per the original Inceptionv3 network (FID provided in the supplement are computed using MI neural estimator and not the Inceptionv3 network). DDPM NLL and WGAN w-distance. \todo{make it formal}
% 
% For a set of generated samples from a method which is larger in size~(1000) than the original sample size, we randomly select samples from the set as many as the real samples to obtain FID score. Having performed 100 trials, we obtain mean and standard deviation of FID scores for a given method. Our method obtains the lowest FID scores w.r.t. all the other methods.
% 
}
% 
% \vspace{-2.0mm}
% 
\label{tab:supplement_eval_metrics}
\end{table}

\begin{figure*}
\centering
\begin{subfigure}{0.33\textwidth}
\includegraphics[width=.99\linewidth]{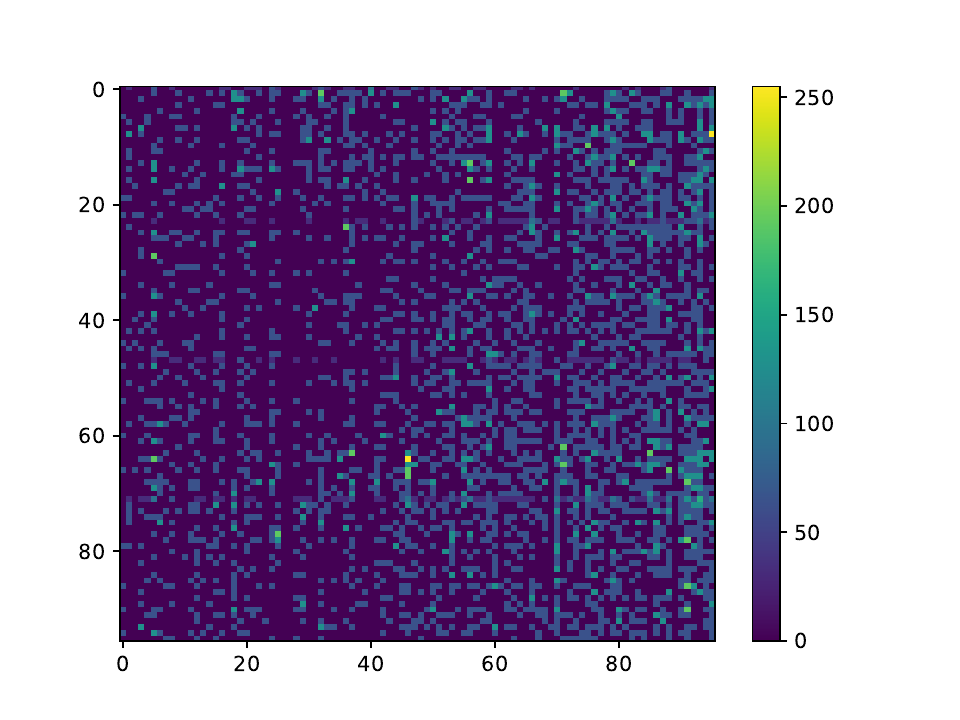}
\end{subfigure}
\hspace{-5mm}
\begin{subfigure}{0.33\textwidth}
\includegraphics[width=.99\linewidth]{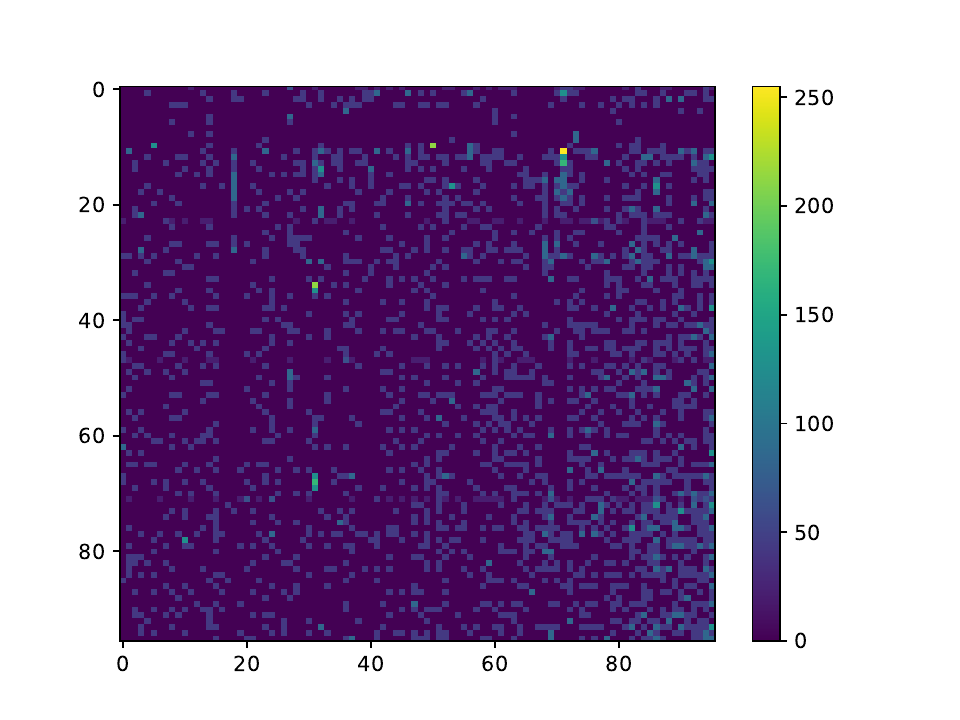}
\end{subfigure}
\hspace{-5mm}
\begin{subfigure}{0.33\textwidth}
\includegraphics[width=.99\linewidth]{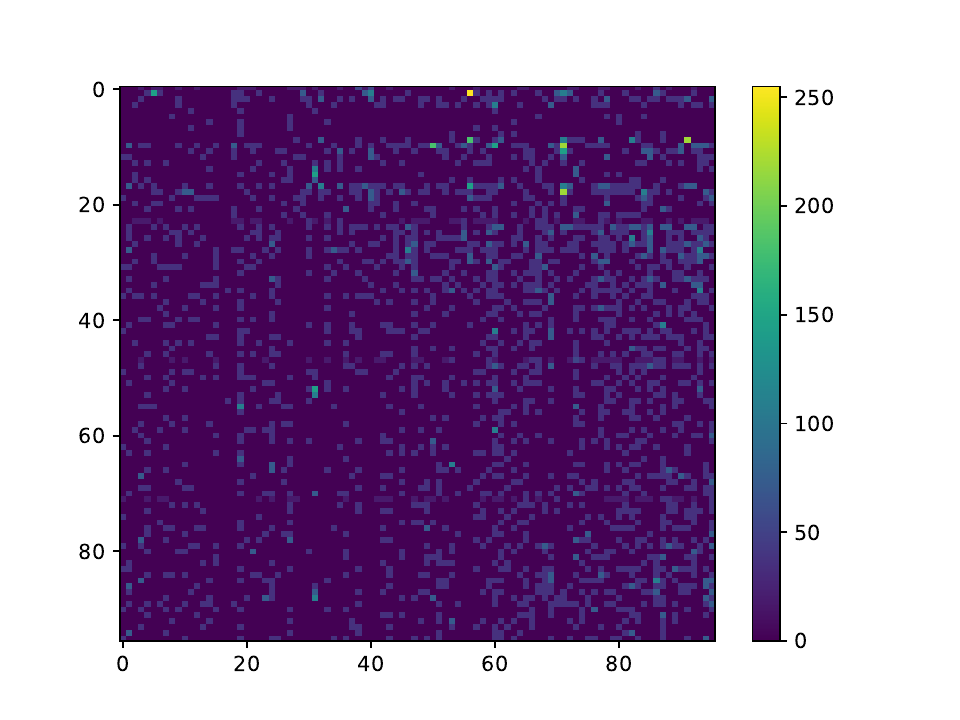}
\end{subfigure}
\begin{subfigure}{0.33\textwidth}
\includegraphics[width=.99\linewidth]{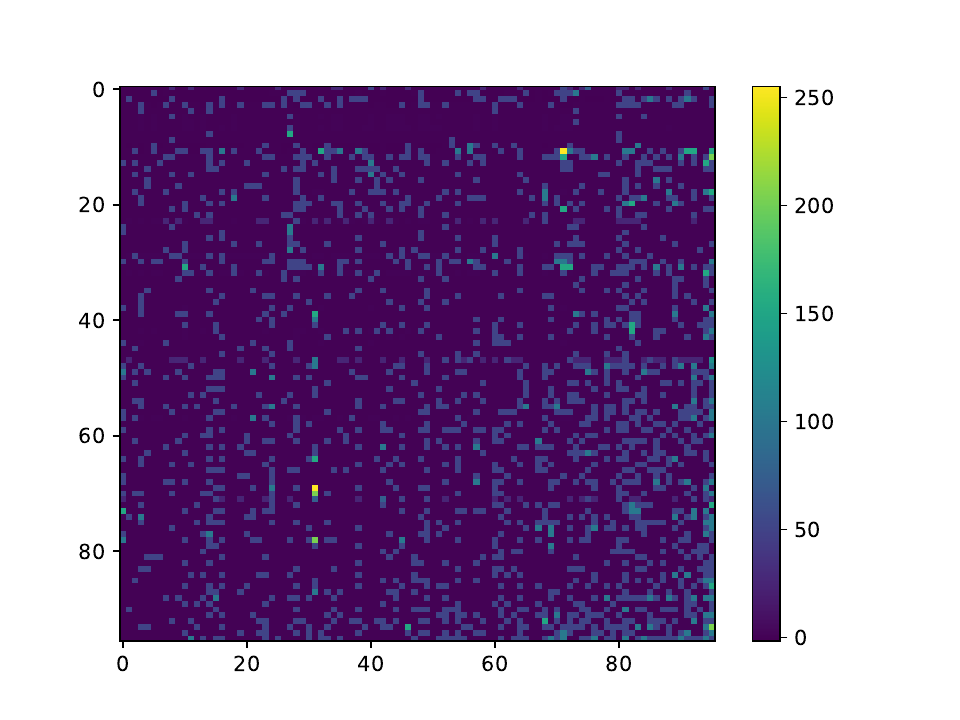}
\end{subfigure}
\hspace{-5mm}
\begin{subfigure}{0.33\textwidth}
\includegraphics[width=.99\linewidth]{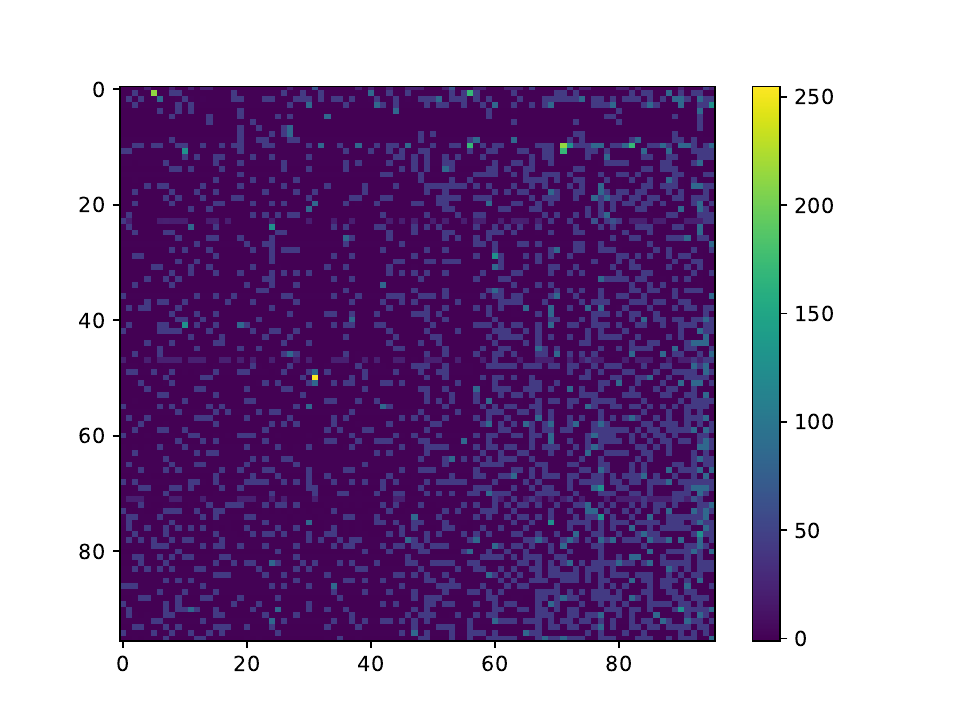}
\end{subfigure}
\hspace{-5mm}
\begin{subfigure}{0.33\textwidth}
\includegraphics[width=.99\linewidth]{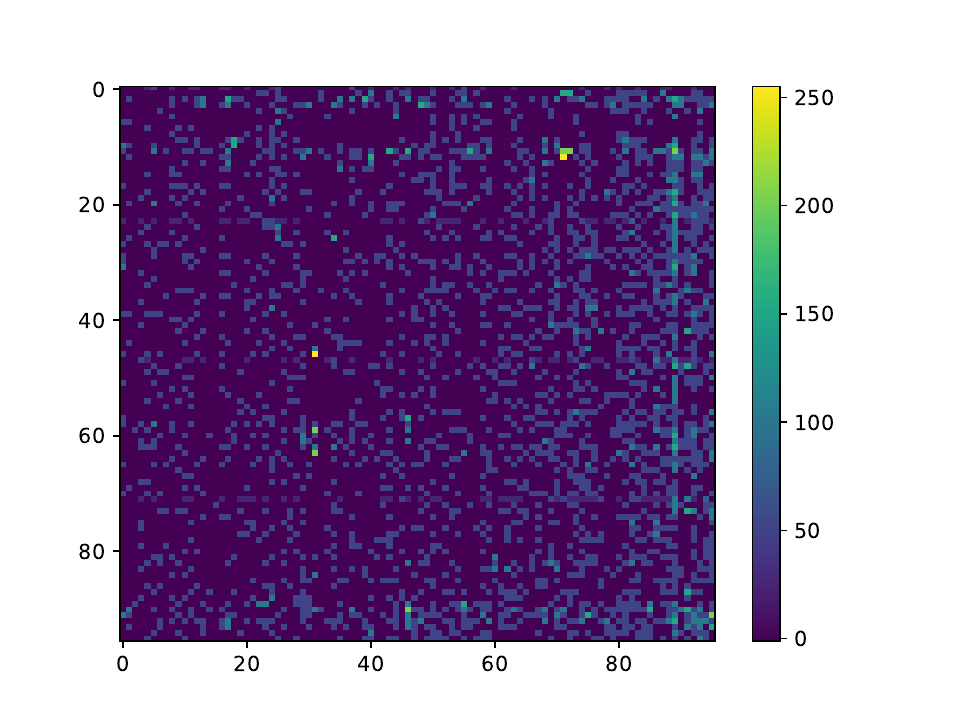}
\end{subfigure}
\caption{Randomly selected images from Neuropixels datasets. The images in the first row are original images whereas the ones in the second row are generated by our approach.}
\label{fig:org_images_neuro}
\end{figure*}

\begin{figure*}
\centering
\begin{subfigure}{0.33\textwidth}
\includegraphics[width=.99\linewidth]{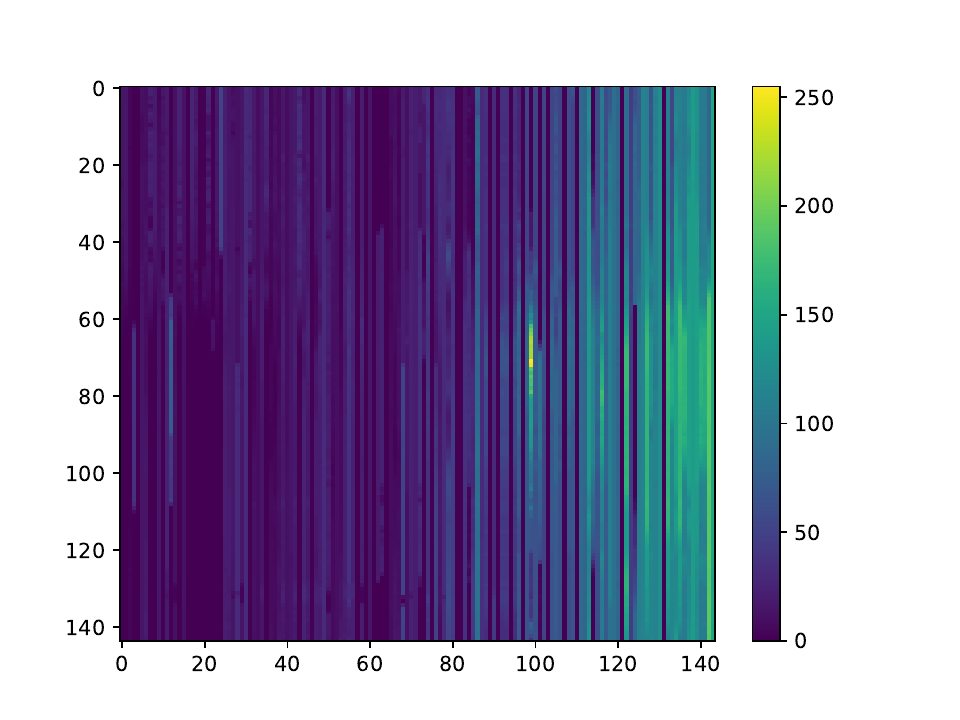}
\end{subfigure}
\hspace{-5mm}
\begin{subfigure}{0.33\textwidth}
\includegraphics[width=.99\linewidth]{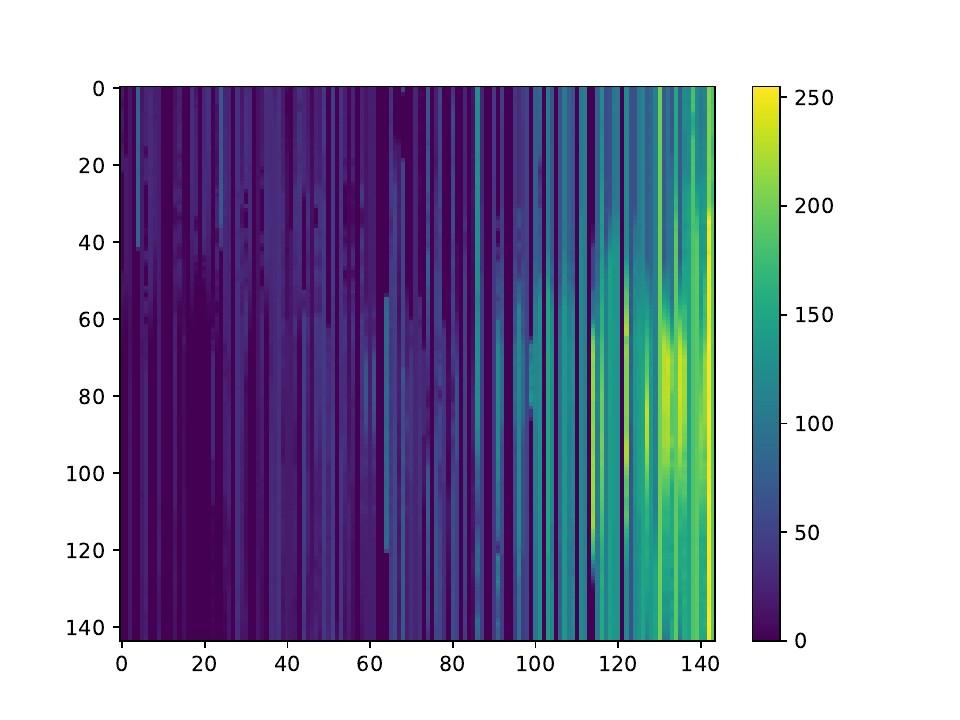}
\end{subfigure}
\hspace{-5mm}
\begin{subfigure}{0.33\textwidth}
\includegraphics[width=.99\linewidth]{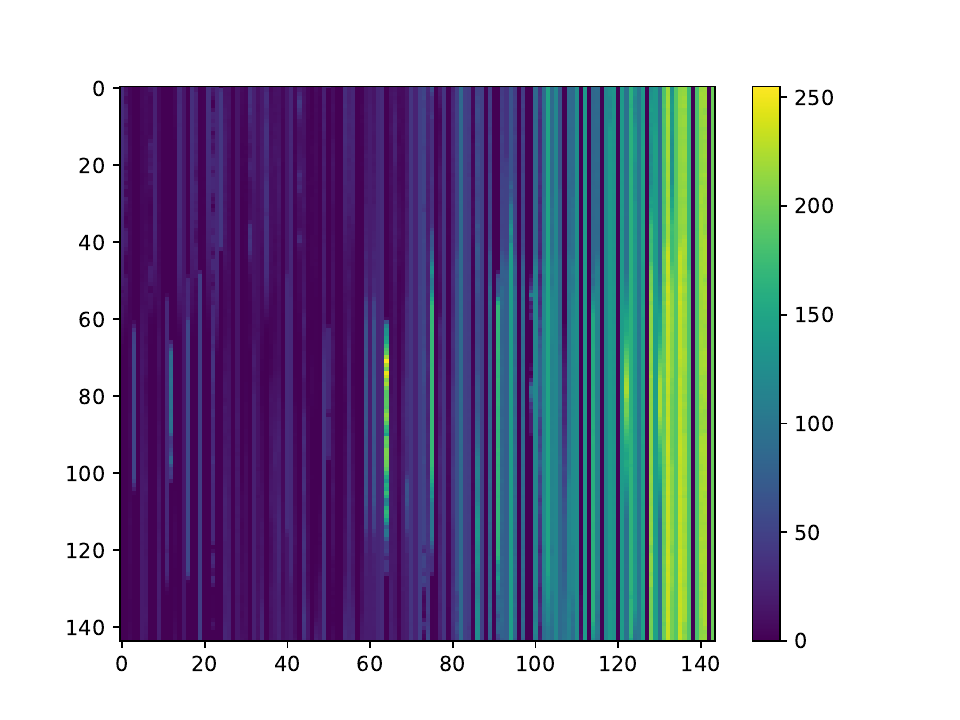}
\end{subfigure}
\begin{subfigure}{0.33\textwidth}
\includegraphics[width=.99\linewidth]{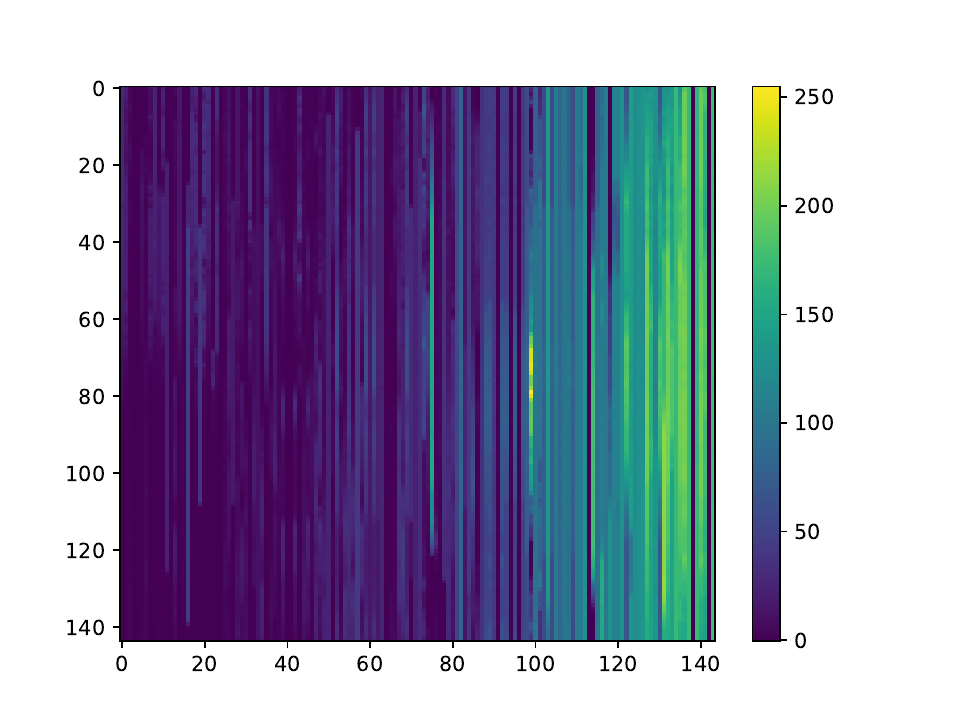}
\end{subfigure}
\hspace{-5mm}
\begin{subfigure}{0.33\textwidth}
\includegraphics[width=.99\linewidth]{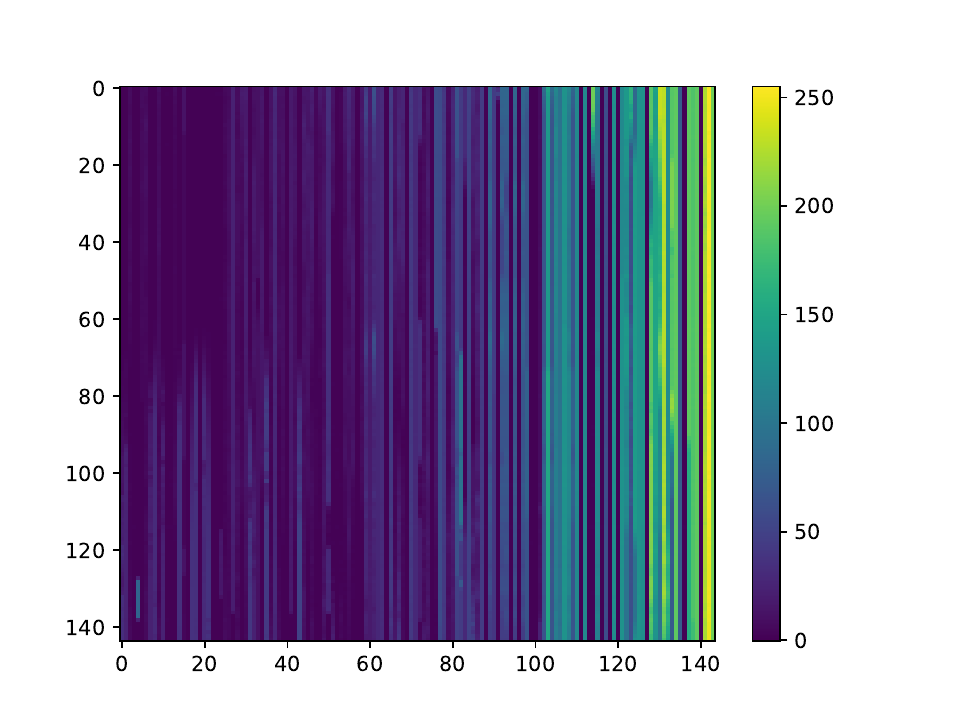}
\end{subfigure}
\hspace{-5mm}
\begin{subfigure}{0.33\textwidth}
\includegraphics[width=.99\linewidth]{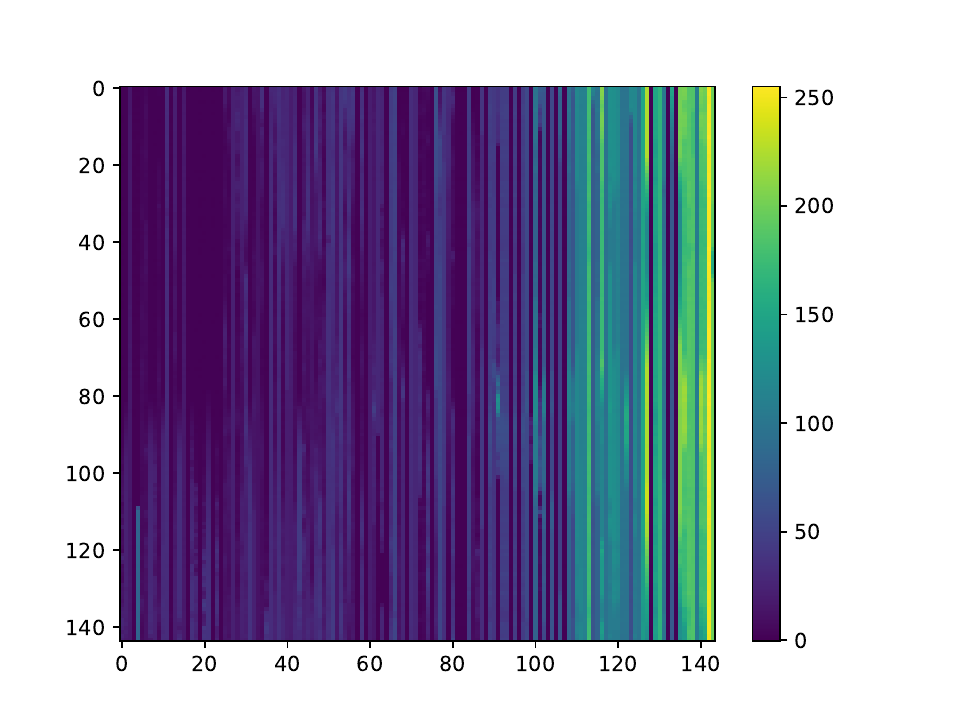}
\end{subfigure}
\caption{Randomly selected images from Wind Farms dataset. The images in the first row are original images whereas the ones in the second row are generated by our approach.}
\label{fig:org_images_wind}
\end{figure*}

\begin{figure*}
\centering
\begin{subfigure}{0.33\textwidth}
\includegraphics[width=.99\linewidth]{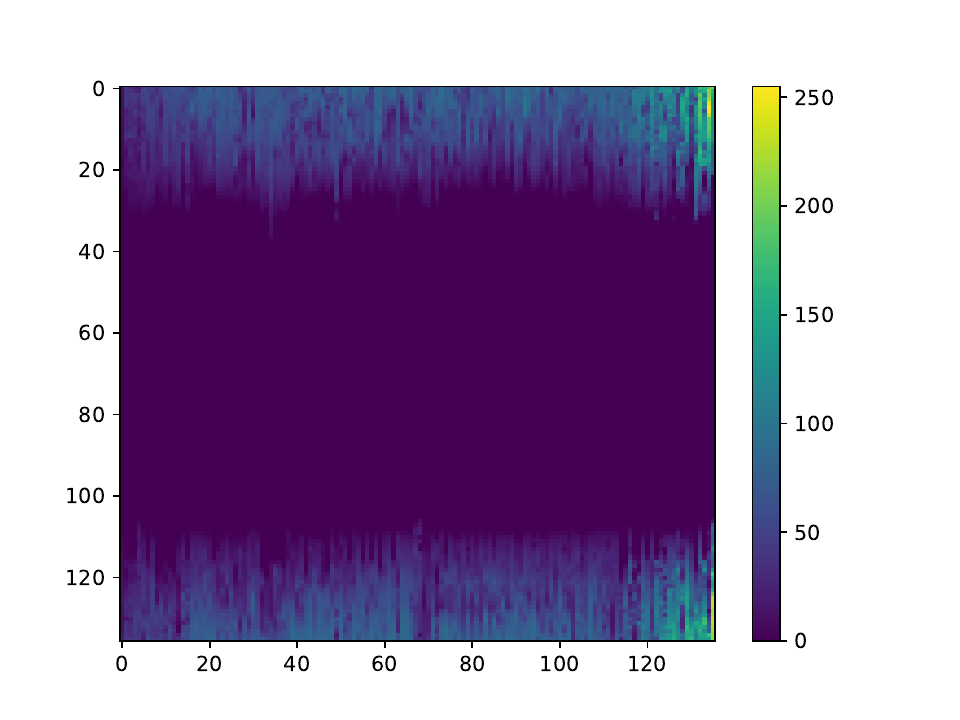}
\end{subfigure}
\hspace{-5mm}
\begin{subfigure}{0.33\textwidth}
\includegraphics[width=.99\linewidth]{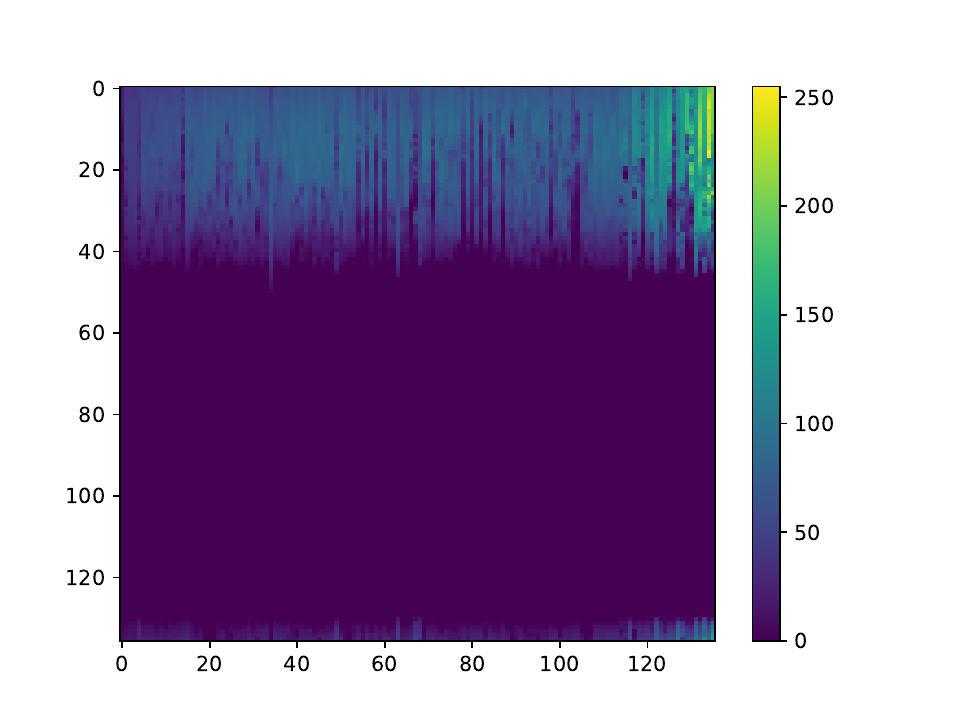}
\end{subfigure}
\hspace{-5mm}
\begin{subfigure}{0.33\textwidth}
\includegraphics[width=.99\linewidth]{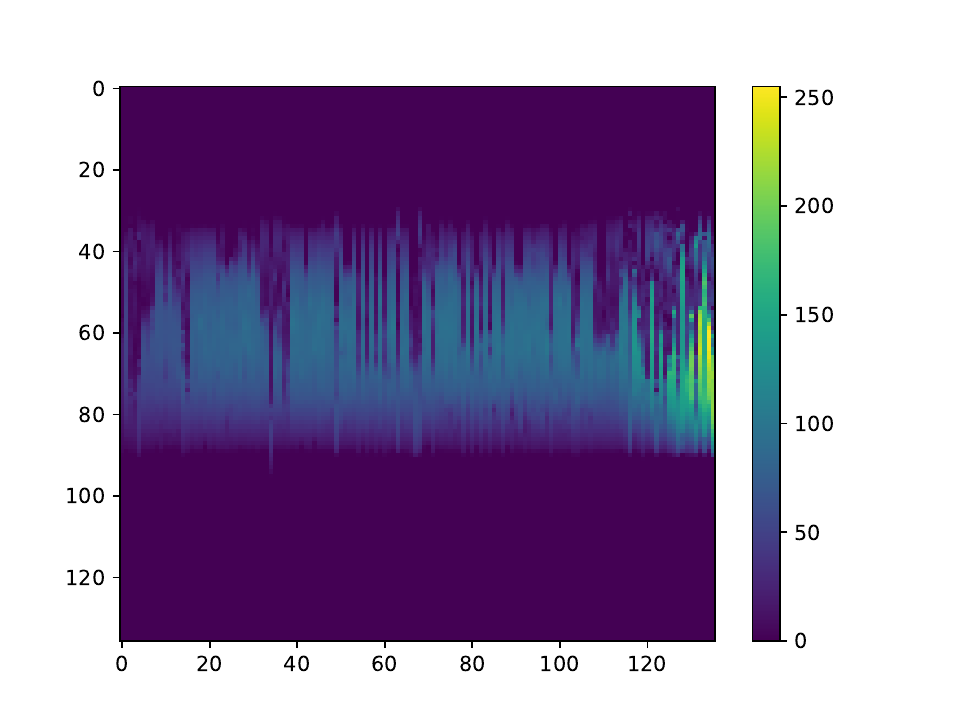}
\end{subfigure}
\begin{subfigure}{0.33\textwidth}
\includegraphics[width=.99\linewidth]{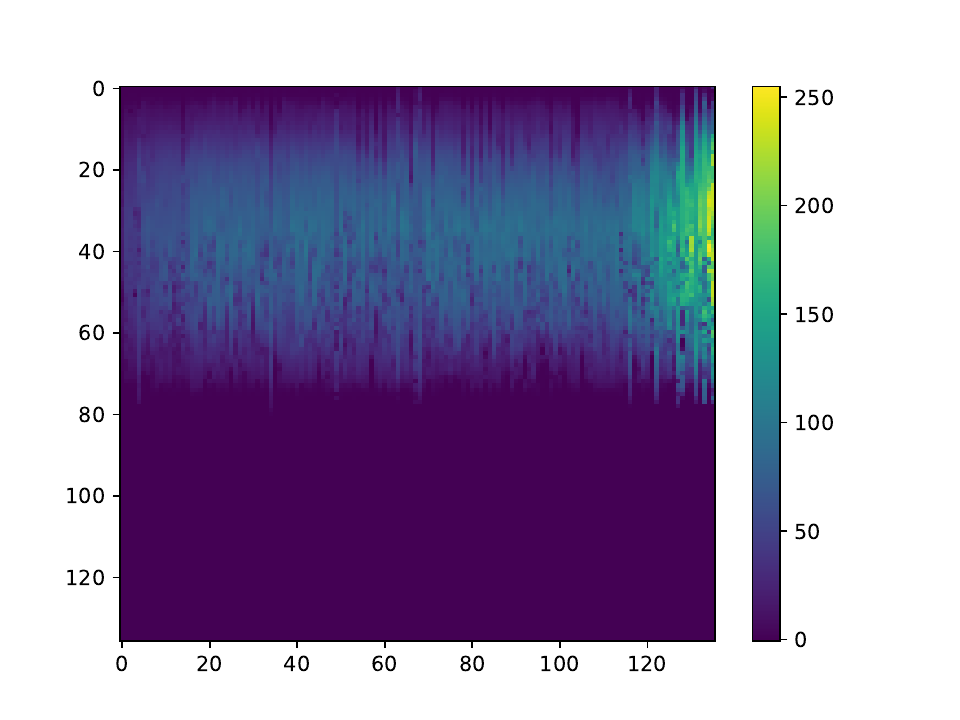}
\end{subfigure}
\hspace{-5mm}
\begin{subfigure}{0.33\textwidth}
\includegraphics[width=.99\linewidth]{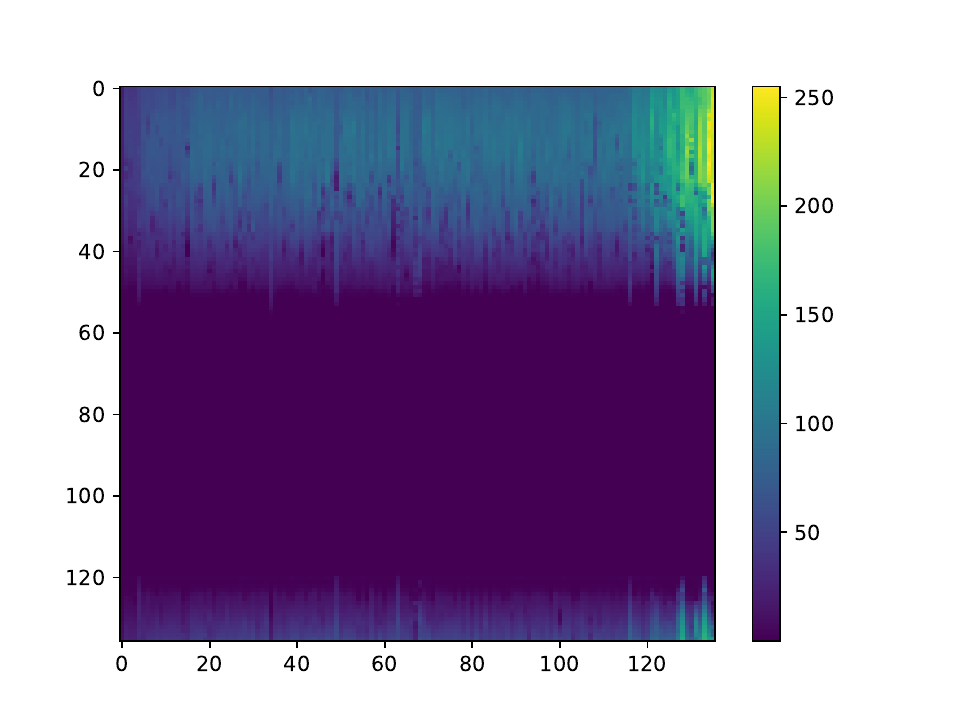}
\end{subfigure}
\hspace{-5mm}
\begin{subfigure}{0.33\textwidth}
\includegraphics[width=.99\linewidth]{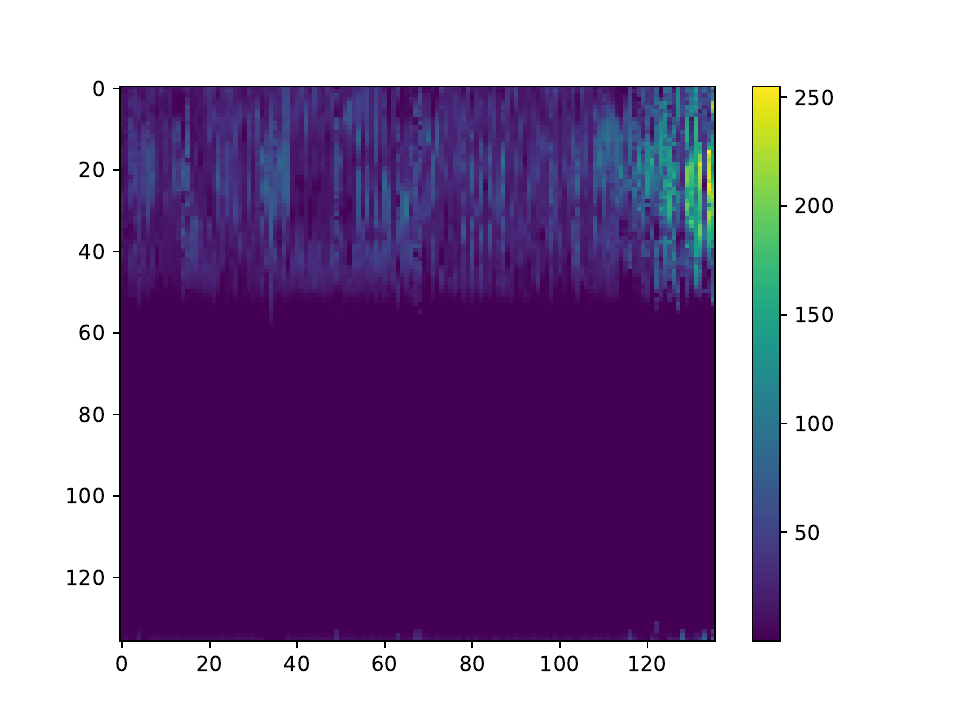}
\end{subfigure}
\caption{Randomly selected images from Solar Power dataset. The images in the first row are original images whereas the ones in the second row are generated by our approach.}
\label{fig:org_images_solar}
\end{figure*}

\paragraph{Visualization of generated samples}
In \figref{fig:org_images_neuro}, \ref{fig:org_images_wind}, and \ref{fig:org_images_solar}, we show visualization of randomly selected images, both from the original dataset as well as the ones generated by our method. Note, while the images are single channel, we present these as colored ones here simply for more appealing visualization as a practitioner from these data domains would do.

\section{Conclusion}

%We introduced the problem of generating multivariate time-series data as an image, which has potential applications in various domains including healthcare. Considering that sample size is super small for such problems, in contrast to datasets of natural images, state-of-the-art deep learning methods for generative sampling are bound to overfit 
We introduced the problem of generating multivariate time-series data in the form of images, with potential applications across diverse domains, including healthcare. Given the inherently small sample sizes in such problems, in contrast to datasets of natural images, state-of-the-art deep learning methods for generative sampling are prone to overfitting. This is due to the fact that in all the methods samples are generated from a canonical distribution and then mapped to data distribution through a highly expressive decoder or denoising diffusion process. To address the issue of sample efficiency, we proposed a deep information theoretic approach for generating samples directly in the dual space of data distribution which is obtained by estimating its divergence in the dual form w.r.t. the respective marginal distribution. Within this framework, we introduced various ideas to ensure robust sampling. Alongside theoretical guarantees, we conducted a comprehensive experimental analysis using several real-world datasets demonstrating the competitiveness of our method in comparison to various established deep generative models. 

\appendix

{
    % \small
    \bibliographystyle{ieeenat_fullname}
    \bibliography{references}
}

\end{document}